\documentclass[preprint]{vgtc}               

\graphicspath{{figures/}{pictures/}{images/}{./}} 

\usepackage{times}                     

\usepackage{tabu}                      
\usepackage{booktabs}                  
\usepackage{lipsum}                    
\usepackage{mwe}                       

\usepackage{mathptmx}                  

\usepackage{amsmath}

\usepackage{caption}
\usepackage{subcaption}

\usepackage{graphicx}
\usepackage[table,xcdraw]{xcolor}

\usepackage{soul}

\onlineid{1508}

\vgtccategory{Research}

\vgtcinsertpkg

\newcommand{\systemname}{OpenFLAME}

\newcommand{\ifhl}[1]{#1}

\definecolor{vps1}{HTML}{B25628}
\definecolor{vps2}{HTML}{4C7C31}
\definecolor{device}{HTML}{6D00C7}

\newcommand{\sagar}[1]{}
\newcommand{\tao}[1]{}
\newcommand{\srini}[1]{}

\title{OpenFLAME: Federated Visual Positioning System \\ to Enable Large-Scale Augmented Reality Applications}

\author{Sagar Bharadwaj\thanks{Equal contribution.}~\thanks{e-mail addresses:\{skalasib, hwillia2, ainiuw, mliang4, taojin, srini, agr\}@andrew.cmu.edu} %
\and Harrison Williams\footnotemark[1]~\footnotemark[2] %
\and Luke Wang\footnotemark[1]~\footnotemark[2] %
\and Michael Liang\footnotemark[2] %
\and Tao Jin\footnotemark[2] %
\and Srinivasan Seshan\footnotemark[2] %
\and Anthony Rowe\footnotemark[2]}
\affiliation{Carnegie Mellon University}

\teaser{
  \centering
  \includegraphics[width=0.9\linewidth]{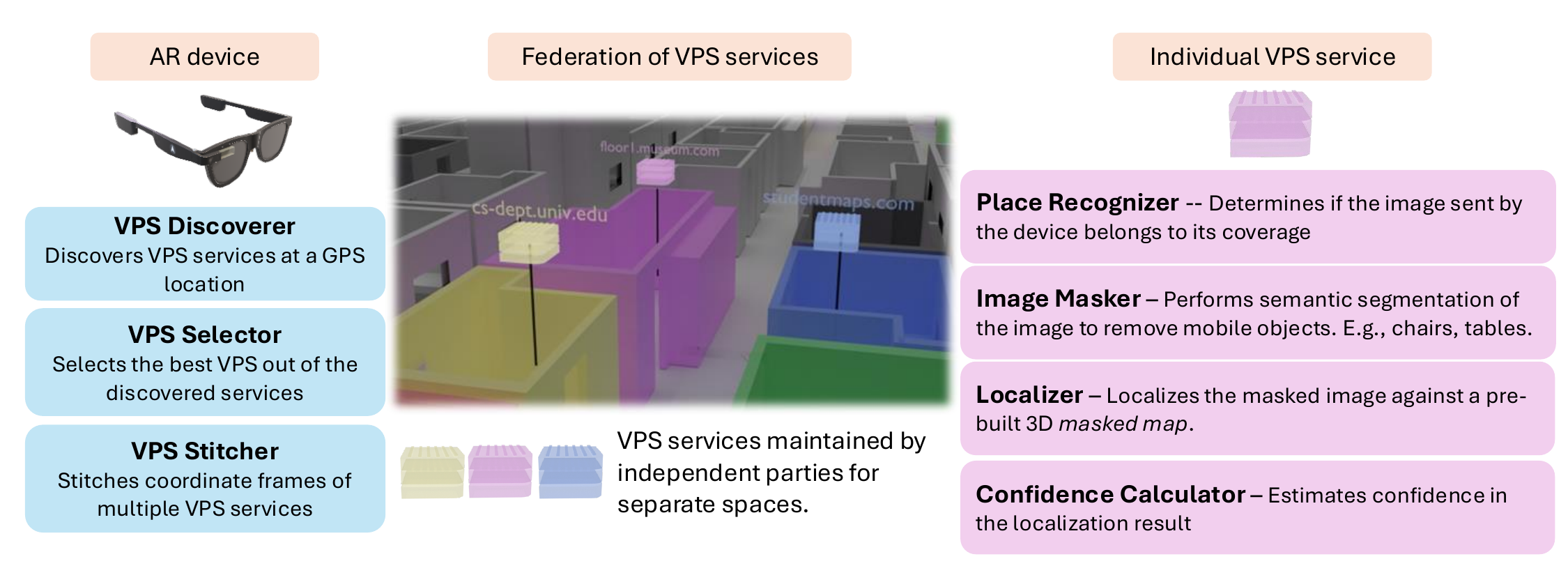}
  \caption{\systemname{} federates the creation and maintenance of Visual Positioning Systems (VPS) allowing independent organizations to support VPS services for their own spaces enabling privacy and distributed maintenance. Its client-side components enable VPS discovery, selection and seamless transition across service boundaries. }
  \label{fig:teaser}
}

\abstract{
    World-scale augmented reality (AR) applications need a ubiquitous 6DoF localization backend to anchor content to the real world consistently across devices. Large organizations such as Google and Niantic are 3D scanning outdoor public spaces in order to build their own Visual Positioning Systems (VPS). These centralized VPS solutions fail to meet the needs of many future AR applications---they do not cover private indoor spaces because of privacy concerns, regulations, and the labor bottleneck of updating and maintaining 3D scans. In this paper, we present \systemname{}, a federated VPS backend that allows independent organizations to 3D scan and maintain a separate VPS service for their own spaces. This enables access control of indoor 3D scans, distributed maintenance of the VPS backend, and encourages larger coverage. Sharding of VPS services introduces several unique challenges---coherency of localization results across spaces, quality control of VPS services, selection of the right VPS service for a location, and many others. We introduce the concept of federated image-based localization and provide reference solutions for managing and merging data across maps without sharing private data.  
} 


\begin{document}


\firstsection{Introduction}

\maketitle

Building world-scale persistent Augmented Reality (AR) applications has been the goal of the community for several decades. Such applications require a ubiquitous, continually updated localization backend ~\cite{jin2024stagear, schonberger2016structure, pan2024glomap, pang2023ubipose} so that 3D content can be anchored anywhere in a continuously changing world. Furthermore, such anchoring must be consistent across devices to enable collaborative AR applications. To support such applications, Visual Positioning Systems (VPS) are being increasingly deployed in public spaces by many organizations (e.g., Google~\cite{geospatial_api}, Niantic~\cite{niantic_lightship}, and Apple~\cite{apple_ar_ahcnor}).

VPS determines the location and orientation of a device with respect to a 3D map using visual features. Existing VPS solutions are usually provided by organizations that have the resources to scan and host vast spans of public spaces. For example, Geospatial API~\cite{geospatial_api}, Google's VPS solution, is based on the 3D world map created by Google's street view imagery collected over 15~years~\cite{google_street_view}. 
These types of VPS solutions fail to meet the expectations of future AR applications in two important ways:

\begin{itemize}
    \item \textbf{Lack of ubiquitous coverage} -- Private spaces would not host their scans on centralized servers due to privacy concerns and lack of fine-grained access control of their data. As a result, existing VPS solutions are limited to public spaces that such organizations can access and scan~\cite{niantic_lightship, geospatial_api}. 
    \item \textbf{Staleness of visual data} -- Real-world scenes evolve over time. Constantly updating scans of dynamic environments requires massive cartography efforts, which are cost-prohibitive for centralized organizations. As a result, the scans get stale, reducing localization accuracy. 
\end{itemize}

A crowdsourced VPS solution accepting contributions from independent parties could perhaps address these limitations~\cite{google_indoor_maps,openstreetmap}. However, not all spaces (e.g., sensitive spaces like private offices and national labs) would consider uploading their maps to crowdsourced VPS services. Furthermore, crowdsourcing all of the data into a handful of large VPS services makes future AR applications reliant on these central organizations. This introduces central points of failure that could bring down a large number of AR applications and concentrates control over future AR applications.

A federated VPS infrastructure that allows independent organizations to collect and host their own scans can overcome these limitations. Federation encourages larger coverage as private data can be independently hosted and access-controlled. It also unlocks the ability to maintain and frequently update a ubiquitous VPS solution in a distributed fashion. \ifhl{Federation enables the system to scale organically, as new VPS services can be added independently without the need for centralized coordination or shared infrastructure.} We present \systemname{}\footnote{OpenFLAME stands for Open Federated Localization and Mapping Engine. \url{https://www.open-flame.com/}}, a federated VPS solution that can enable world-scale 6 Degrees-of-Freedom (DoF) localization across dynamically changing private and public spaces. In \systemname{}, a device that wants to localize itself first discovers VPS services available to it at the given GPS location. The device then contacts each of these VPS services, potentially hosted by independent organizations, and sends visual cues to all of them. Once localization results are received from these VPS services, the device will select the best VPS service for the location. The device also calculates and applies transformations to VPS results to ensure they are coherent across service boundaries. Federation of VPS services introduces some unique challenges absent in today's centralized solutions:

\begin{itemize}
    \item \textbf{Coherent localization across spaces} -- The localization trajectory exposed to AR applications has to be coherent across spaces, even if the individual 3D scans of spaces are in their own separate coordinate frames of reference. To maintain privacy, the VPS system cannot rely on sharing visual features between different VPS services to stitch these disparate coordinate frames. 
    \item \textbf{High variability in service quality} -- The quality of VPS services hosted on \systemname{} could have a wide range from scans made using low-quality RGB images to high-quality lasers. We need a solution for devices to dynamically pick the highest-quality VPS service at a location.
    \item \textbf{Irrelevant queries to services} -- Individual VPS services in \systemname{} might receive localization requests from devices outside of their coverage, resulting in a waste of resources used for localization. We need a lightweight solution to weed out requests that are outside the coverage area of a server before triggering the localization pipeline. 
\end{itemize}

In this paper, we introduce the concept of federated VPS and provide effective solutions to the above challenges. 
To summarize, our contributions are as follows:

\begin{itemize}
    \item We present the design of a distributed federated VPS infrastructure called \systemname{}, where independent organizations maintain their own VPS services for their own spaces. 
    
    \item We present simple solutions to solve challenges unique to federated VPS---a \textit{VPS Stitcher} to enable coherent trajectories across VPS services without exposing visual features, a \textit{VPS Selector} to dynamically select the highest quality VPS service at a location, a \textit{Place Recognizer} to filter out requests outside of a server's VPS coverage, and a \textit{Pose Confidence Calculator} to estimate the confidence of pose estimates. While the solutions to these individual components are borrowed from existing fields of study, their integration enables our vision of a federated VPS backend.
    
    \item At individual VPS services, we present a localization pipeline based on scene understanding that enables the handling of dynamically changing scenes.
\end{itemize}

We evaluate each of the proposed components in dynamically changing large-scale indoor environments in \S~\ref{sec:evaluation}. We implement an AR 3D indoor navigation application on top of \systemname{} to show that it can support the development of large-scale AR applications. The implementations of the VPS pipeline, AR application, client-side library, and supplementary tools will be open-sourced after publication.

\section{Example Workflow}

Consider an AR cultural heritage guided tour application that displays infographics overlaid on real-world objects indoors and outdoors. For example, outdoors, it might show fact bubbles and timelines of landmark historical buildings, while indoors, it might describe museum artifacts and paintings. To accurately anchor 3D content against the real world, such an application would need a localization solution that works across indoor and outdoor spaces while seamlessly transitioning between environments.

Existing VPS solutions, mostly limited to public areas, cannot support such applications. Even if some museums have 3D-scanned their floors, no convenient system today can integrate these scans with outdoor 3D localization systems (e.g., Google Geospatial APIs) to provide coherent VPS solutions to applications. While the museum could request one of the large providers to integrate their scans with the larger system, it would deter the museum from implementing fine-grained access control to their scans (e.g., only people with tickets can view and localize against museum scans). Furthermore, the process of updating the map would be bottlenecked at the large VPS provider.

In \systemname{}, the VPS services for specific regions are maintained by independent organizations. In our example, the museums would maintain their own 3D scans, and the localization service would be provided against those scans. When outdoors, the AR application would use a large VPS provider, such as Google, to localize and anchor 3D content. Once the user enters the museum, \systemname{} will discover and switch to the VPS service maintained by the museum. The application would rely on \systemname{} to discover VPS services, select the best VPS service for a location, get 6DoF poses, and stitch results from multiple VPS services.

Furthermore, individual VPS services in \systemname{} implement a pipeline to support dynamically changing environments. We use simple image segmentation to remove frequently moving objects from the scan and visual cues while localizing. For example, the pipeline allows furniture and people to be removed from museum scans while retaining the predominantly stationary artifacts, increasing the scan's longevity and relevance.
\section{Background}

\subsection{Visual positioning}

Visual Positioning Systems aim to estimate the 6 DoF pose of a device camera using visual input. Early VPS methods often rely on visual markers such as AprilTags~\cite{olson2011apriltag} and ArUco~\cite{aruco}, where known 2D templates with pre-defined geometry are detected in the image and mapped to corresponding 3D points. This enables 6-DoF pose estimation by solving the Perspective-n-Point (PnP) problem~\cite{li2012robust}. However, deploying visual markers throughout a space is often impractical. Hence, extensive work has been focusing on markerless visual localization using hand-crafted feature descriptors such as SIFT\cite{lowe2004distinctive}, SURF\cite{bay2006surf}, and ORB\cite{rublee2011orb} to detect and describe visual features. These features are typically matched against a database of 3D points constructed via Structure-from-Motion \cite{schonberger2016structure} or Simultaneous Localization and Mapping \cite{mur2015orb}. Once 2D-3D feature correspondences are established, the camera pose can be estimated by solving the PnP problem. Learning-based methods have recently improved the localization pipeline, with keypoint detectors and descriptors like SuperPoint \cite{detone2018superpoint}, and matchers such as SuperGlue \cite{sarlin2020superglue} and LightGlue \cite{lindenberger2023lightglue} boosting reliability and accuracy. 

\subsection{VPS services with ubiquitous coverage}

VPS services with ubiquitous coverage aim to localize users in expansive environments such as city streets, parks, and commercial venues. Unlike indoor VPS, which can rely solely on visual input due to its relatively small search space, large-scale VPS typically adopts a hybrid approach---using GPS and IMU data to obtain a coarse initial 6 DoF pose estimate, which is then refined through visual observations captured by the device’s camera. Several commercial systems have successfully implemented this approach. Google’s Geospatial API \cite{geospatial_api} allows users to determine precise device poses by combining GPS data with visual localization against Google Street View imagery as a global reference. Similarly, Apple's ARKit ARGeoAnchor \cite{ar_geoanchor} also leverages location and camera input to anchor AR content in real-world locations. Niantic Lightship \cite{niantic_lightship} has taken a crowdsourced approach by asking users to scan public spaces at various times of day and from multiple viewpoints, building a rich and diverse database that enables robust localization under varying lighting conditions. These large-scale VPS services demonstrate the feasibility of global visual localization, enabling persistent AR experiences and navigation assistance in complex real-world environments. 



\section{Characteristics of individual VPS services}

Figure~\ref{fig:indoorOutdoorScans} shows two 3D scans from independent organizations---Google (\ref{fig:outdoorScan}) and a private university (\ref{fig:indoorScan}). The two scans, and as a result, the VPS services provided on top of these scans, vary in multiple ways, such as reconstruction quality (which in turn affects pose estimation accuracy) and the coordinate system used. 
In this section, we explore the characteristics of individual VPS services and how they differ from one another.

\begin{figure}
    \centering
    \begin{subfigure}{.2\textwidth}
        \centering
        \includegraphics[width=.95\linewidth]{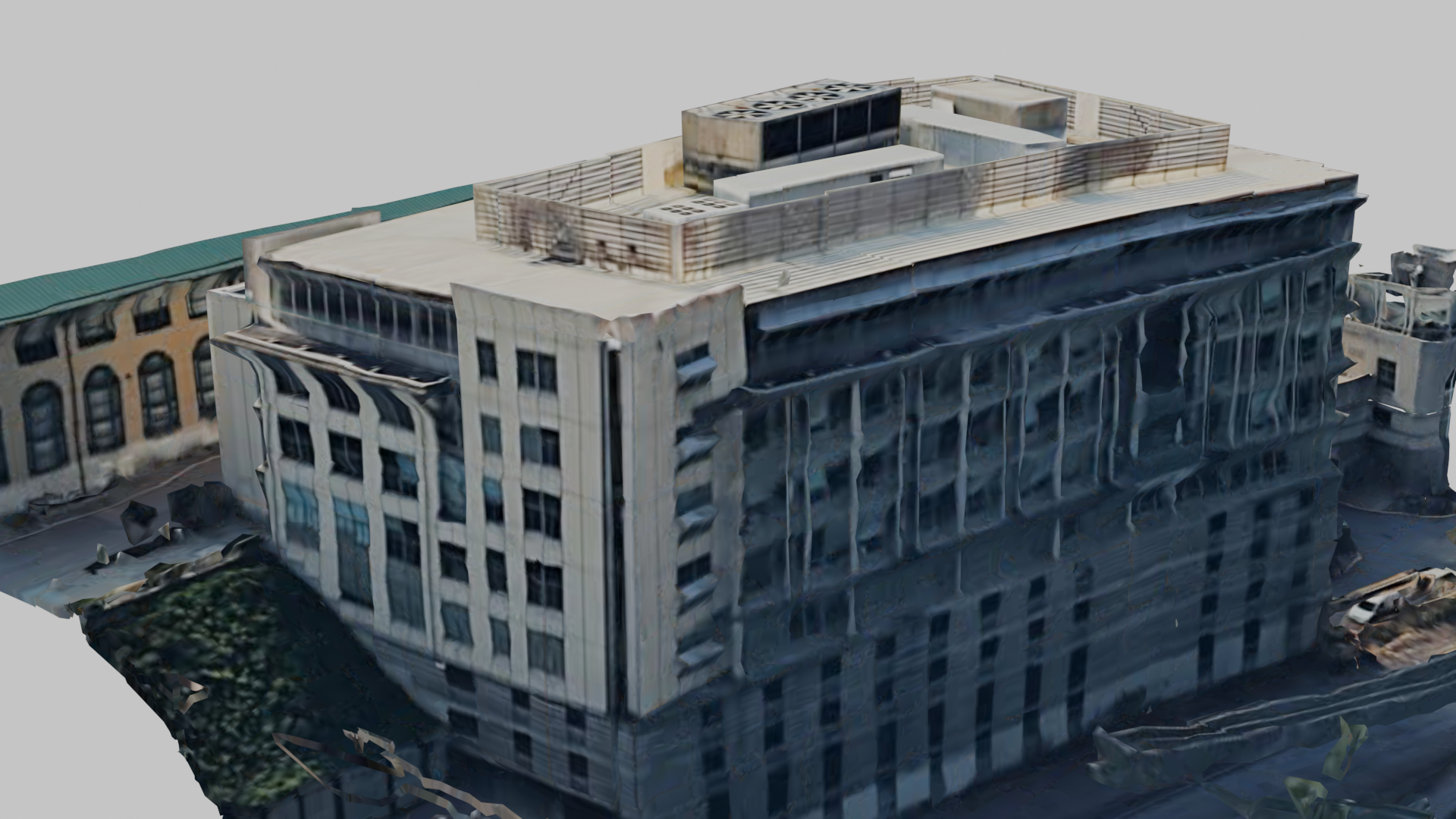}  
        \caption{Outdoor scan by Google.}
        \label{fig:outdoorScan}
    \end{subfigure}
    \hfill
    \begin{subfigure}{.2\textwidth}
        \centering
        \includegraphics[width=.95\linewidth]{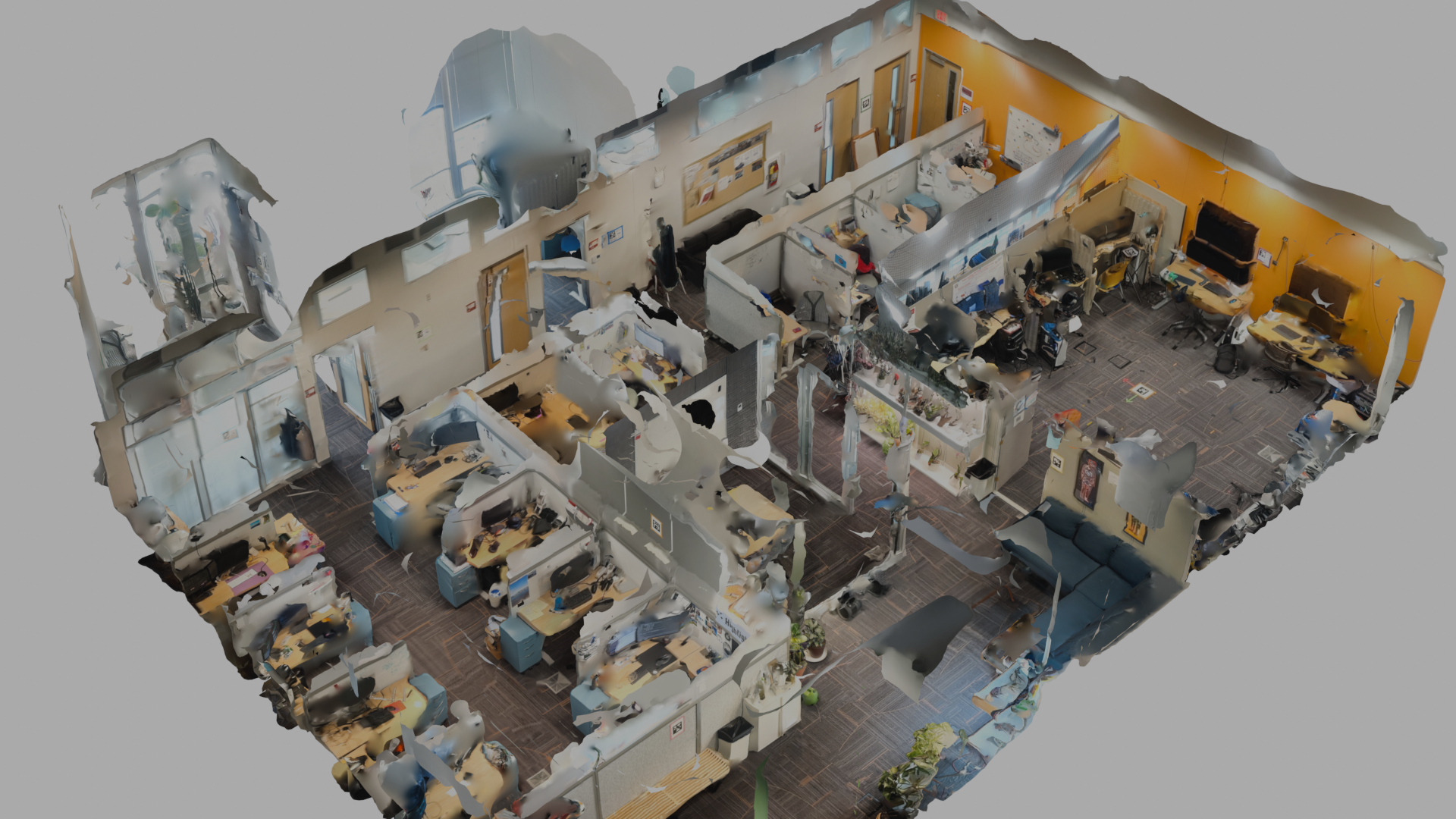}  
        \caption{Indoor scan by a university.}
        \label{fig:indoorScan}
    \end{subfigure}
    \hfill
    \begin{subfigure}{.3\textwidth}
        \centering
        \includegraphics[width=.95\linewidth]{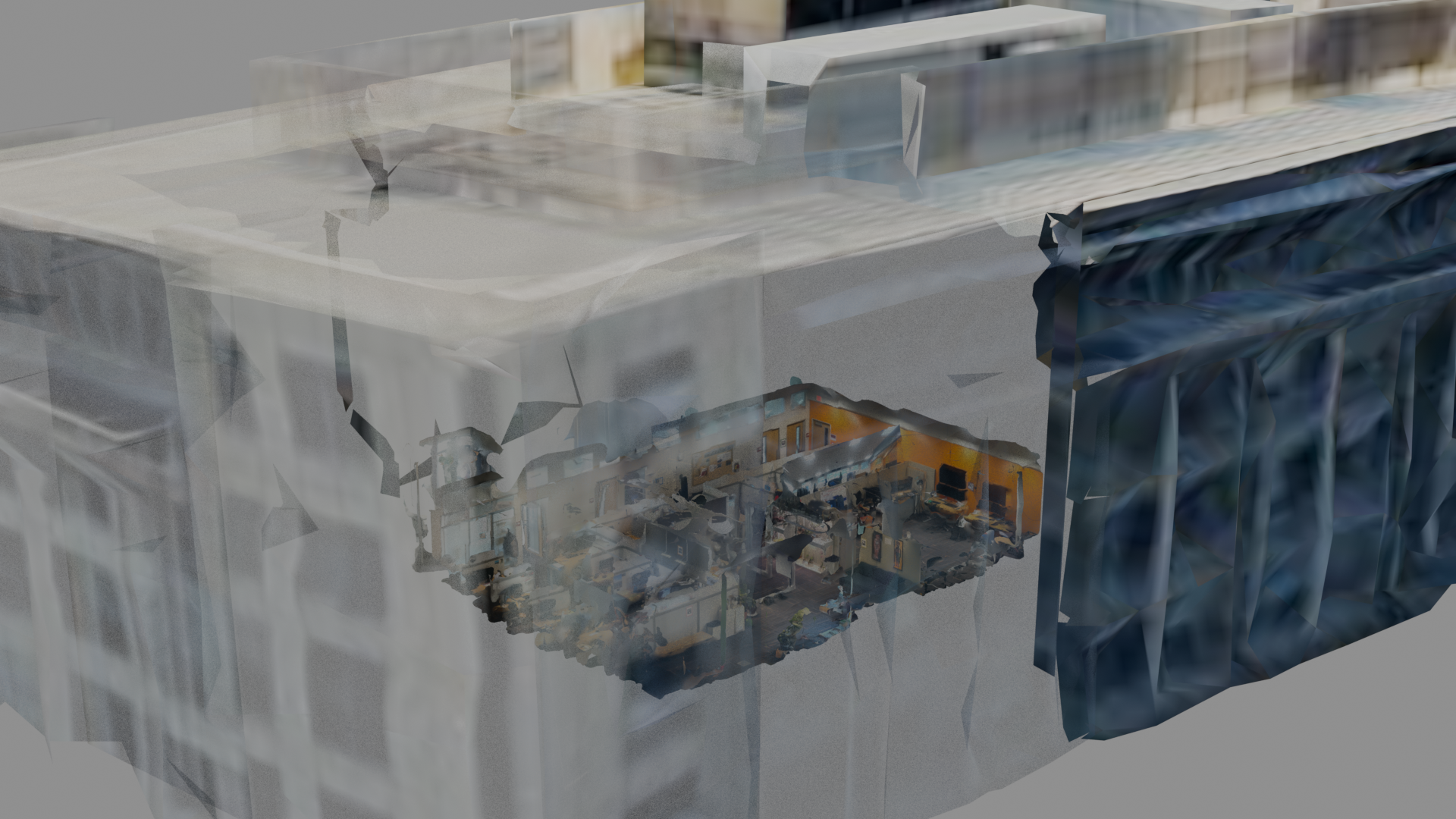}  
        \caption{\systemname{} integrates multiple VPS services.}
        \label{fig:indoorOutdoorScan}
    \end{subfigure}
    \caption{\ifhl{\mbox{\systemname{}} integrates independent VPS services.}}
    \label{fig:indoorOutdoorScans}
\end{figure}

\textbf{Quality} -- \systemname{} supports VPS services of varying qualities. For example, in Figure~\ref{fig:outdoorScan}, Google's scan is created with data from high-quality cameras and LiDARs mounted on Google Street View cars and aerial imagery. In contrast, the university's indoor scan in Figure~\ref{fig:indoorScan} is created using the Polycam application~\cite{polycam} on an iPhone. Admitting VPS services of varying quality encourages incremental deployment of VPS without expecting perfection from the start, thereby increasing VPS coverage. 

\textbf{Disparate coordinate systems} -- \systemname{} allows individual 3D scans to be in their own coordinate systems and does not require alignment with the global geographic coordinate system. Precise alignment with latitude, longitude, and altitude requires expensive survey equipment such as Total Stations~\cite{total_station} and RTK GNSS~\cite{rtk_gnss}, increasing the barrier to deployment. For example, the university indoor scan in \ref{fig:indoorScan} is in its own coordinate system, unlike Google's scan, which is laid out in the system of latitudes and longitudes. 

\textbf{Overlap of VPS coverages} -- We do not enforce spatial exclusivity---different VPS services can cover the same area. 

\section{Localization pipeline}
\label{sec:lolPipeline}

\begin{figure}[htp]
    \centering
    \includegraphics[width=1.0\linewidth]{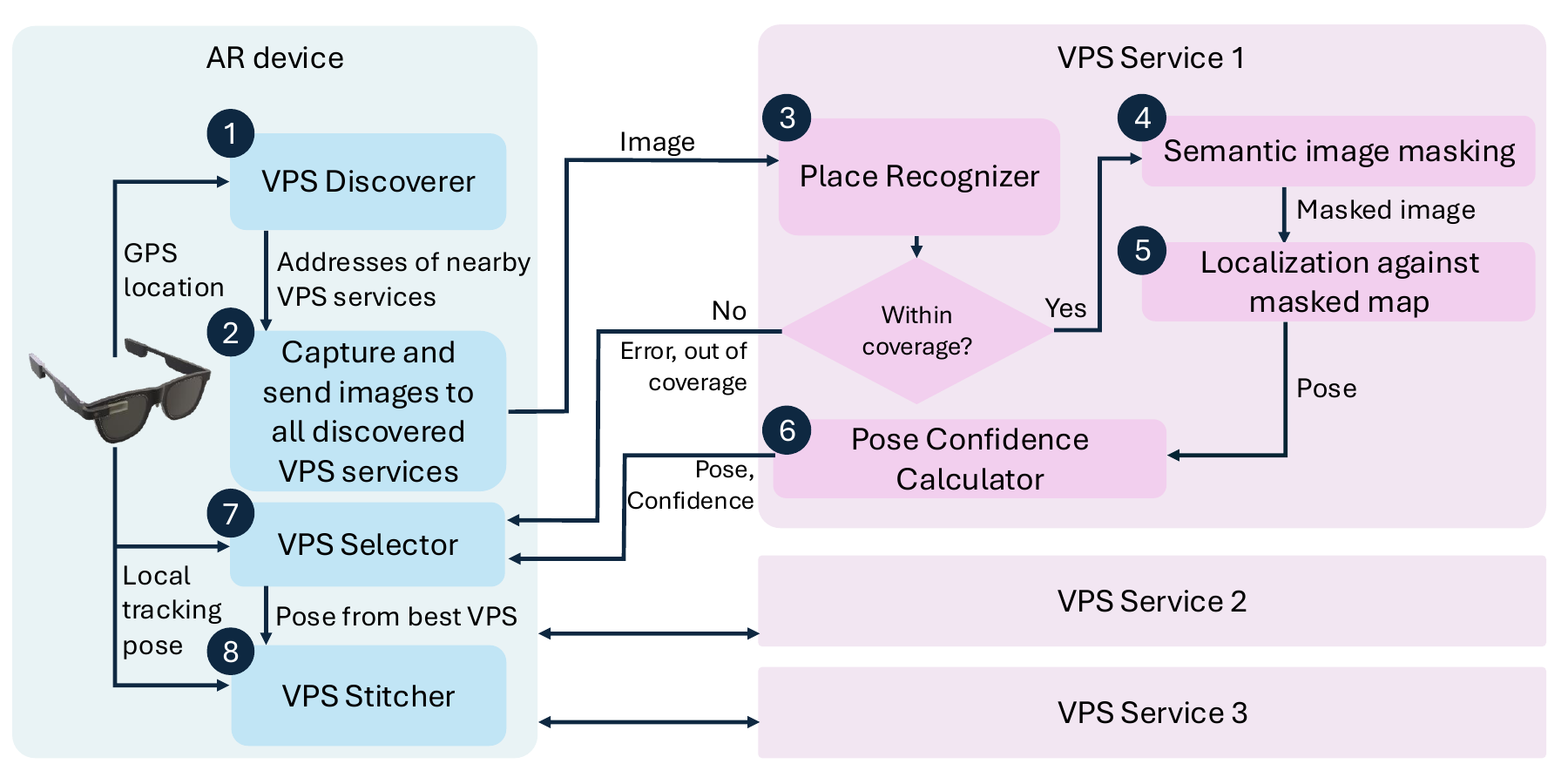}
    \caption{Localization pipeline on \systemname{}.}
    \label{fig:locPipeline}
\end{figure}

Figure~\ref{fig:locPipeline} shows the localization pipeline of \systemname{}. The device first uses \systemname{}'s \textit{VPS Discoverer} module to identify all VPS services offered at the device location. The device then sends visual cues to a select subset of these services. Individual \systemname{} VPS services perform image-based localization while ignoring dynamic parts of the map, such as furniture and people. Individual VPS services return the estimated pose and confidence scores to the device. The device then selects the best localization results and transforms it to stitch it with the coordinate system of previous pose estimates from other VPS services. 

The localization pipeline shown in Figure~\ref{fig:locPipeline} is run periodically at a pre-configured interval (1 second by default). Results from local tracking libraries (e.g., ARCore~\cite{ar_core}, ARKit~\cite{ar_kit}, and WebXR~\cite{web_xr}) are used to anchor 3D elements in the intermediate frames. Furthermore, we do not trigger the entire localization pipeline every cycle. Since devices are not expected to frequently cross VPS service boundaries, we assume that a selected VPS service can be used for multiple consecutive localization cycles. The device continues to use the current service until the pose confidence scores drop; at this point, a new discovery query is triggered to identify a more suitable service. The remainder of this section describes in detail each of the components in Figure~\ref{fig:locPipeline}. 

\subsection{VPS Discoverer}


The VPS Discoverer module is responsible for identifying the VPS services available at a given GPS location. To accomplish this, it requires access to data that maps geographic regions to corresponding lists of VPS services. This data can be stored and queried in various ways using existing systems, including spatial databases (e.g., GeoFire~\cite{geo_fire}, PostGIS~\cite{PostGIS}, MongoDB~\cite{mongodb}), Geographic Information Systems (GIS) such as ArcGIS~\cite{arc_gis} or Carto~\cite{carto}, or even through DNS-based approaches~\cite{gibb2022spatial, bharUniting, gibb2023earth}. 
In our implementation, we repurpose the Domain Name System (DNS) to function as the VPS discoverer. This approach offers several advantages: it leverages widely available infrastructure, supports caching mechanisms, is straightforward to implement, and naturally enables federation. In this paper, we specifically focus on the challenges of image-based localization in a federated setting. A detailed treatment of the organization and querying of spatial data for VPS service discovery is out of scope of this paper.

\subsection{Capture and send images}

Once the VPS services at a given location are discovered, \systemname{} requests the device to capture an image. It then broadcasts the image to a subset of the discovered VPS services. The filtering mechanisms used to select a subset of VPS services are configurable by the AR applications using \systemname{}. For example, the application can whitelist a subset of Top Level Domains (TLDs) that are acceptable. An AR campus navigation application, for instance, can choose to use only VPS services hosted on \texttt{.edu} domains.
The application can also choose to limit the number of VPS services that are being contacted every discovery cycle, in which case a subset is arbitrarily chosen. After a few discovery and localization cycles, \systemname{} eventually locks in on the VPS service that is the most accurate for a location using the \textit{VPS Selector} module described later. 
\systemname{} provides parameter settings that the applications can use to configure the method of filtering VPS services. 

\subsection{Place Recognizer}
\label{sec:placeRecognizer}

Unlike centralized VPS services that have to satisfy all localization requests from devices, the VPS services on \systemname{} might receive requests from devices outside of their VPS coverage. This is especially true indoors, where GPS errors are high, and the device might discover and make requests to multiple maps before narrowing down its location to a single map. In \systemname{}, VPS services include the Place Recognizer module that takes in an image and determines if it belongs to a place that is within the VPS coverage of the server. This is run before triggering the localization pipeline to conserve system resources.

The Place recognizer module is primarily based on the CLIP~\cite{clip} model---a neural network trained on a large dataset of image-text pairs. We specifically use the image encoder layer of CLIP, which converts an image to an \textit{image embedding} (i.e., a vector), expected to represent the semantic information in the image. To ensure discriminability between spatially close and visually similar spaces (e.g., office areas used by different groups in a university), we fine-tune CLIP for specific VPS services. We construct a dataset of images from such neighboring locations. We modify the original CLIP model by appending a projection head—a fully connected layer—that maps CLIP’s high-dimensional image embeddings to a new 256-dimensional embedding space. During training, we freeze CLIP’s parameters and train only the projection layer using a triplet loss. This encourages embeddings of images from the same room to have lower cosine distance, while pushing apart embeddings from different rooms. The result is a model that produces more discriminative representations suited for fine-grained place recognition. While such fine-tuning enhances performance in areas where data sharing between VPS services is possible, we observe that even the pre-trained CLIP model---without any fine-tuning---performs well in settings where cross-service data access may be restricted.

A CLIP image embedding data set is constructed offline for all the images in the 3D map database---the \textit{database embeddings}. To determine if a given query image lies within the VPS service's coverage, we first get the CLIP embedding of the query image. The query embedding is compared to the database embeddings to find the one with the smallest cosine distance. If the smallest cosine distance exceeds a pre-configured threshold, the image is discarded, and the rest of the localization pipeline is skipped.  The threshold can be changed based on how resource-conserving the VPS service wants to be. At low thresholds, most images are discarded. If the minimum Euclidean distance is high, the VPS service responds with an error code that indicates to the device that the queried image is out of coverage. \S~\ref{sec:eval:placeRecognizer} evaluates the performance of the Place Recognizer module. 

\subsection{Semantic image masking}

\begin{figure}
    \centering
    \includegraphics[width=0.95\linewidth]{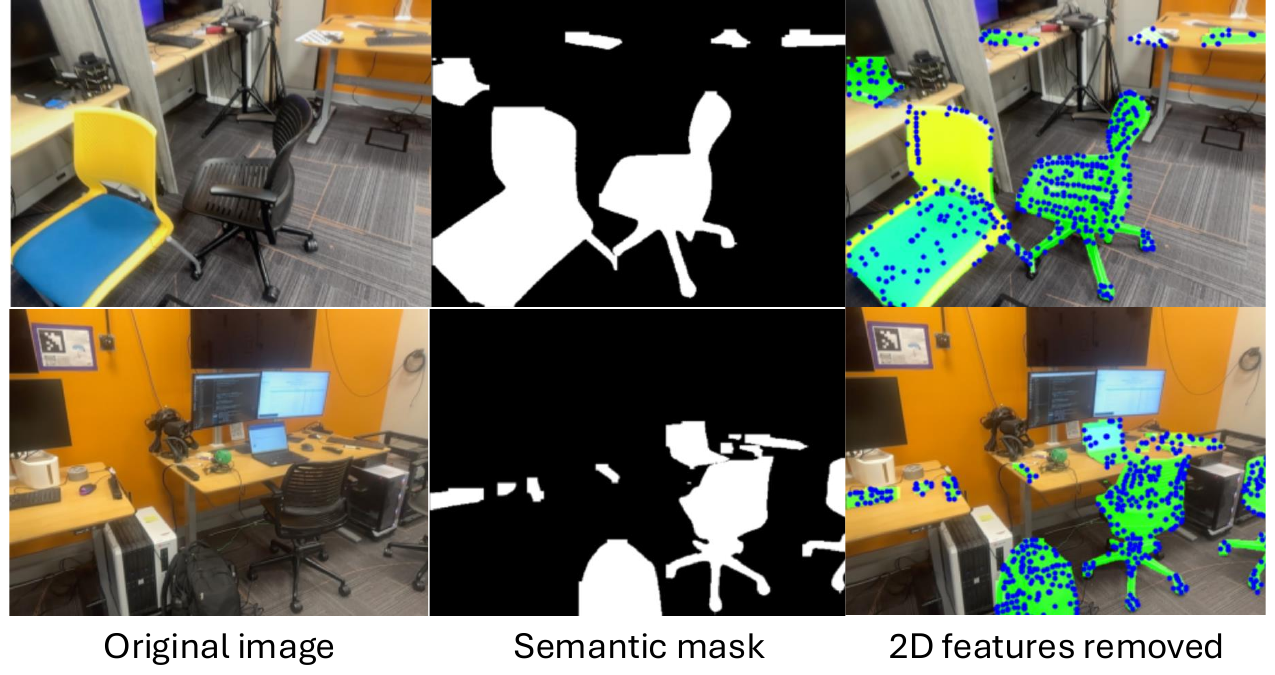}
    \caption{\ifhl{Semantic image masking is used to ignore features from typically dynamic objects, making localization more robust to changes in the environment.}}
    \label{fig:semanticImageMasking}
\end{figure}

Individual VPS services might be hosted in dynamically changing spaces with many objects that frequently change their positions, such as chairs, keyboards, backpacks, and people. As visual positioning is sensitive to the position of visual features, the localization quality drastically reduces because of the frequent motion of such objects. In our pipeline, we address this issue by masking out classes of objects that frequently move in an environment and not considering them in pose estimation.

Figure~\ref{fig:semanticImageMasking} shows two examples of semantic masking. The left column shows the original image. To generate the \textit{semantic mask} in the middle column, we use YOLOv8~\cite{yolo}---a real-time object detection system. The classes of objects to be detected are configurable and can be set by individual VPS services. The figure shows a typical university office environment, and objects such as chairs, keyboards, and backpacks are detected. In a different environment, a different set of object classes can be used. Once 2D image features are detected in the \textit{pose estimation} step (\S~\ref{sec:PoseEstimation}), the features inside the boundaries of the detected objects are removed and are not considered in the rest of the localization pipeline. The rightmost column in Figure~\ref{fig:semanticImageMasking} shows the set of 2D features that are removed. \S~\ref{sec:eval:maskedLocalization} evaluates the benefits of masking dynamic features.

\subsection{Pose estimation}
\label{sec:PoseEstimation}

We use \texttt{hloc} (Hierarchical Localization)~\cite{sarlin2019coarse} to estimate the pose of the given query image against a 3D map. The process of constructing the masked 3D map is described in \S~\ref{sec:scanning}. \texttt{hloc} uses SuperPoint~\cite{detone2018superpoint} feature detector and descriptor, and SuperGlue~\cite{sarlin2020superglue} feature matcher. The combination of learned feature detector, descriptor, and matcher makes the pose estimation robust to lighting changes. To support large maps, hloc also adds a \textit{global retrieval} layer that uses NetVLAD~\cite{arandjelovic2016netvlad} to isolate pose estimation to a small portion of the map that is relevant to the query image. 

We optimize the implementation of hloc to integrate well with \systemname{}. Specifically, the default implementation of the localization pipeline on hloc loads the neural network weights of all models (i.e., SuperPoint, SuperGlue, and NetVLAD) into memory for every localization request. We persist the weights in memory across multiple localization queries to speed up pose estimation. Furthermore, the implementation was not written with support for concurrent executions. We made modifications to the way the query image is loaded to enable concurrent pose estimations. While our implementation uses hloc, this component can be replaced by any vision-based localization method, including the recent synthesize-and-localize~\cite{Sandström2023ICCV, rosinol2023nerf, keetha2024splatam, zhu2022nice} methods. We build the rest of the components in \systemname{} to be agnostic to the method used for pose estimation, which enables VPS services to take advantage of the rapid advances in the field of vision-based localization.

\subsection{Pose Confidence Calculator}
\label{sec:poseConfidence}

Individual VPS services return a confidence score along with their pose estimates to enable the devices to make an informed selection and switch to a different VPS service once the device moves outside of the coverage area. The \textit{Pose Confidence Calculator} module helps estimate the confidence of the computed pose.

In image-based localization, common pose confidence metrics include the number of feature inliers~\cite{fischler1981random}, reprojection error~\cite{reproj_error}, and the ratio of inliers to detected keypoints. However, these feature-based metrics are tightly coupled to the specific pose estimation method and cannot be reliably compared across different VPS services. This coupling would also limit the ability of VPS services to independently adopt new localization techniques, such as emerging synthesize-and-localize methods~\cite{rosinol2023nerf, keetha2024splatam}, which do not rely on feature matching. Additionally, feature-based metrics can fail in indoor environments with repetitive textures (e.g., floor tiles, carpets), where incorrect poses may still yield high inlier counts. To address these limitations, we instead use image similarity between the query image and a rendered view from the 3D scan, enabling consistent and technology-agnostic confidence estimation across VPS services.

\begin{figure}
    \centering
    \includegraphics[width=0.99\linewidth]{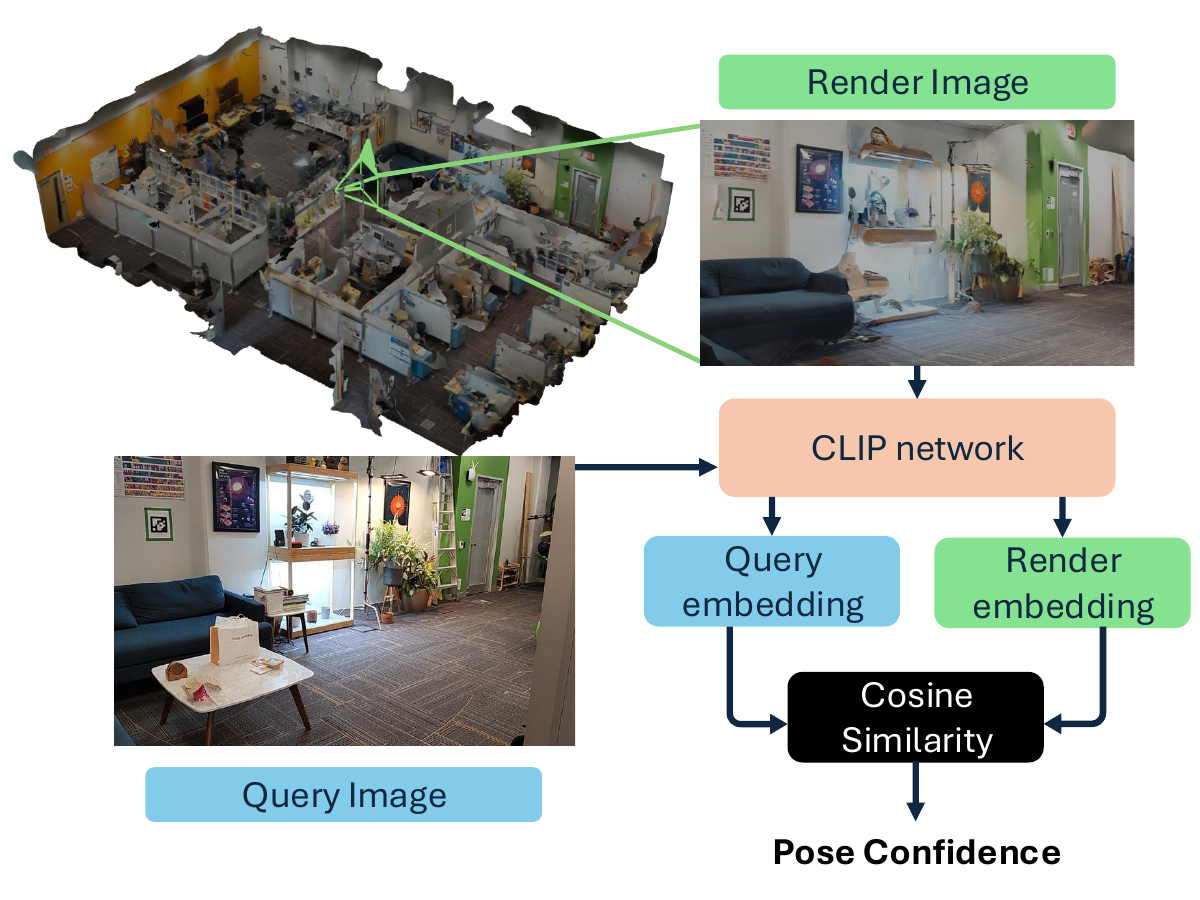}
    \caption{\ifhl{The Pose Confidence component estimates how reliable the calculated pose is by comparing the query image with one rendered from that pose.}}
    \label{fig:poseConfidence}
\end{figure}

Figure~\ref{fig:poseConfidence} shows our pipeline to estimate pose confidence. Once we estimate the pose of the query image, we render the corresponding pose from our 3D scan at the estimated pose. CLIP embeddings for both the rendered image and the query image are generated. The cosine similarity between the embeddings of the rendered and query images is returned as pose confidence. Our pose confidence scores are independent of repeating textures and the number of features as CLIP compares semantic differences between the images. To improve performance, we extend CLIP with a projection layer and fine-tune it so that the final embeddings of the query image and the 3D render at the estimated pose are close in the embedding space. As in \S~\ref{sec:placeRecognizer}, we train this model using triplet loss on a dataset of query and rendered image pairs. In \S~\ref{sec:eval:poseConf}, we justify our choice of using a fine-tuned CLIP model by comparing it to other similarity metrics.

\subsection{VPS Selector}
\label{sec:VPSSelector}

Once the device has all pose estimations from all the VPS services that it sent images to, it has to select one VPS service that is the most accurate for the current location. 
If all the VPS services implement the pose confidence module (\S~\ref{sec:poseConfidence}), and the device trusts these VPS services, it can use these scores for its selection (e.g., a museum AR application would trust the VPS service hosted by the museum administrators and use its pose confidence scores). However, in our federated setting, not all VPS services can be trusted. Some services might return high confidence---either accidentally or with malicious intent---to entice devices to continue using them. Some might only return pose estimates without an attached confidence score. Therefore, devices need a VPS selection method that is independent of the server estimated confidence. If the device consistently finds discrepancies between the device-calculated score and the VPS service returned score, it can mark the VPS service as undesirable and avoid it in future localization cycles.

\begin{figure}
    \centering
    \includegraphics[width=0.99\linewidth]{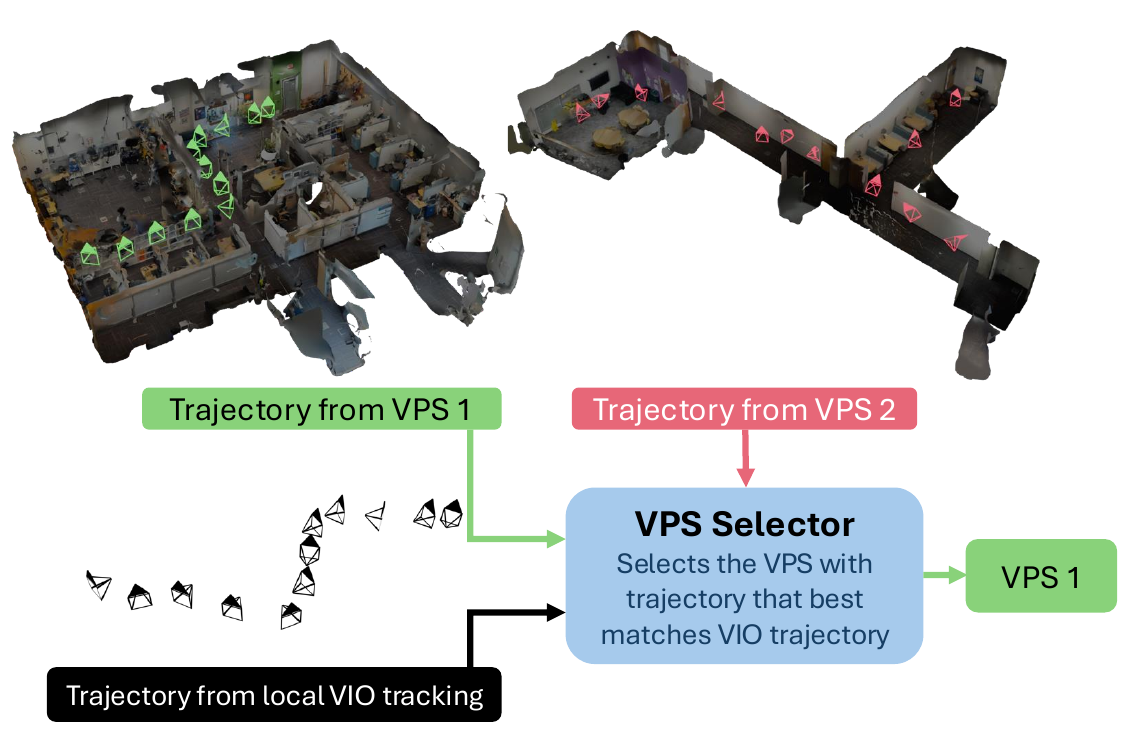}
    \caption{VPS Selector \ifhl{chooses the best VPS service by comparing different VPS trajectory results against on-device VIO tracking.}}
    \label{fig:vpsSelector}
\end{figure}

Figure~\ref{fig:vpsSelector} shows our server-independent VPS selection method. This method can work after the device has captured and collected pose estimates for at least 3 images. The trajectories estimated by all the VPS services are first aligned with the local device trajectory calculated using VIO (Visual Inertial Odometry) algorithms. The trajectories need alignment, as local tracking and tracking with the VPS services each run in their own coordinate systems. Once trajectories are aligned, we calculate the Absolute Trajectory Error (ATE) between server-provided and on-device trajectories. The VPS service with the least trajectory error is selected. \S~\ref{sec:eval:vpsSelector} shows that this technique selects the correct VPS service in almost all cases.

\subsection{Visual features-free dynamic VPS Stitcher}
\label{sec:dynStitching}

Once the device obtains a pose from the selected VPS service, it must align this result with poses from other VPS services encountered in previous cycles, as each service operates in its own coordinate frame. Although stitching 3D scans using visual features at region boundaries is a well-studied problem~\cite{rizk2020real, tang2021map}, we cannot rely on such feature sharing between VPS services due to privacy constraints and pose estimation technology variations.

\begin{figure}[!htp]
    \centering
    \includegraphics[width=0.95\linewidth]{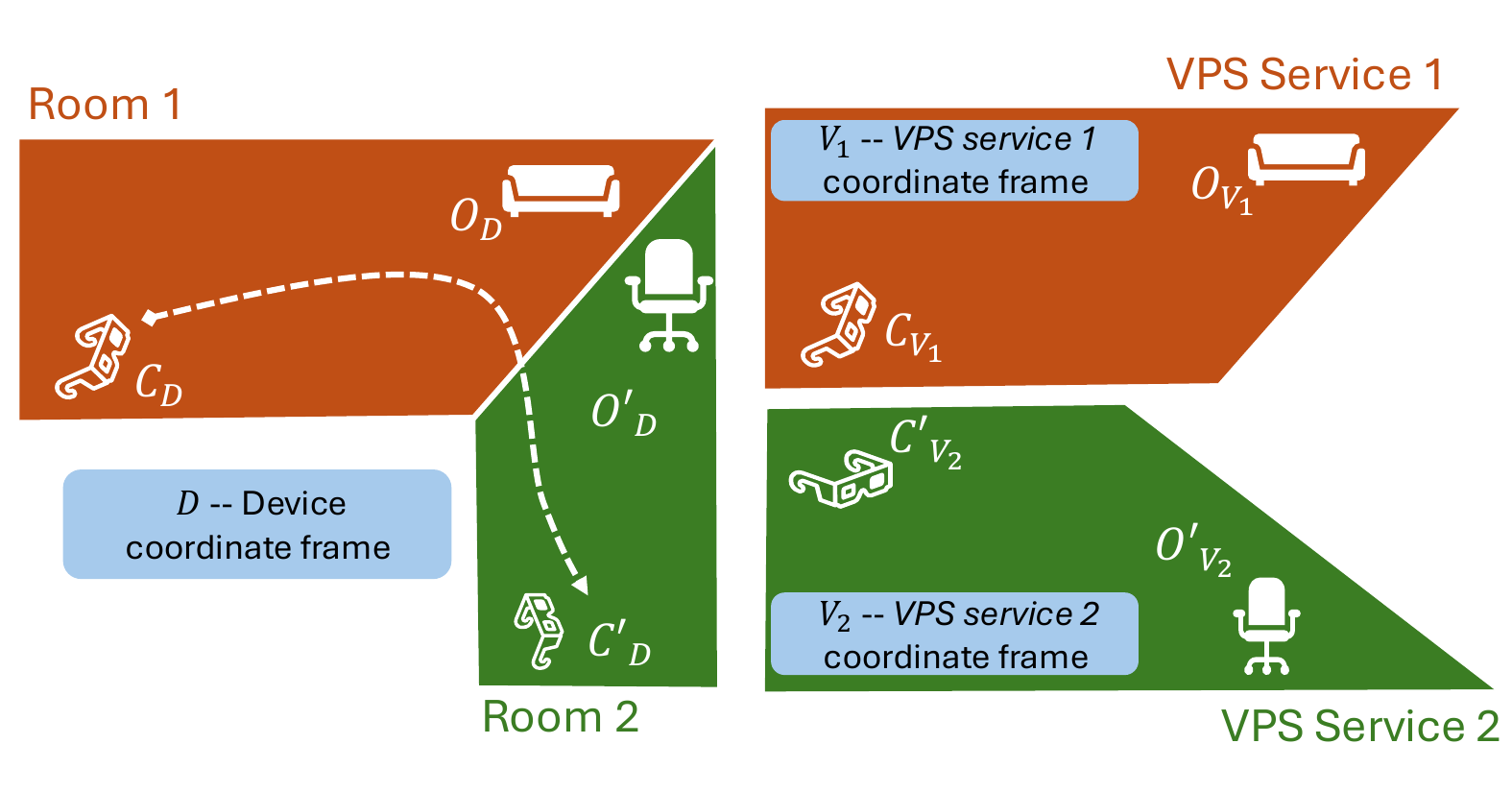}
    \caption{Dynamically stitching coordinate systems.}
    \label{fig:dynamicStitching}
\end{figure}

To understand how \systemname{} dynamically stitches coordinate systems together without requiring visual features, consider two rooms, \textcolor{vps1}{Room 1} and \textcolor{vps2}{Room 2} shown in Figure~\ref{fig:dynamicStitching}. Room 1 and 2 are covered by different VPS services (\textcolor{vps1}{VPS 1} and \textcolor{vps2}{VPS 2}). The two VPS services have their own scans of the respective rooms in their own coordinate systems---\textcolor{vps1}{$V_1$} and \textcolor{vps2}{$V_2$} (shown on the right side of Figure~\ref{fig:dynamicStitching}.) A device running an AR application is also building its own local coordinate system, \textcolor{device}{$D$}. For example, such a coordinate system is built by ARCore~\cite{ar_core} in Android applications, ARKit~\cite{ar_kit} in Apple devices, or WebXR~\cite{web_xr} in web applications. The left side of Figure~\ref{fig:dynamicStitching} shows the layout of the rooms in the real world and the trajectory of the device determined by local AR tracking in the coordinate frame \textcolor{device}{$D$}.

Consider an object $O$ in \textcolor{vps1}{Room 1}. Let the $4 X 4$ pose matrix of the object in the \textcolor{vps1}{$V_1$} coordinate frame be \textcolor{vps1}{$O_{V_1}$}. Similarly, the pose matrix of the same object $O$ in the local coordinate frame \textcolor{device}{$D$} is \textcolor{device}{$O_D$}. A device camera moves from Room 1 to Room 2 as shown on the left side of Figure~\ref{fig:dynamicStitching}. The initial position of the device, as calculated in the local tracking coordinate frame, is \textcolor{device}{$C_D$}. The device also sends a localization request to VPS server 1 to get its pose in the \textcolor{vps1}{$V_1$} coordinate frame, \textcolor{vps1}{$C_{V_1}$}. The pose of the object $O$ with respect to the camera is the same irrespective of which coordinate system is considered. Therefore, we have:

\begin{equation}
     \textcolor{device}{C_D^{-1} \, O_D} = \textcolor{vps1}{C_{V_1}^{-1} \, O_{V_1}} \\
    \implies \textcolor{device}{O_D} = \textcolor{device}{C_D} \, \textcolor{vps1}{C_{V_1}^{-1} \, O_{V_1}}
\end{equation}

Let the final position of the device camera in Room 2, in the local tracking coordinate frame \textcolor{device}{$D$}, be \textcolor{device}{$C'_D$}. The pose of the camera as calculated by VPS service 2, in the \textcolor{vps2}{$V_2$} coordinate frame is \textcolor{vps2}{$C_{V_2}$}. Following the same argument as above, for a different object $O'$ in Room 2, we have:

\begin{equation}
     \textcolor{device}{C_D'^{-1} \, O'_D} = \textcolor{vps2}{C_{V_2}'^{-1} \, O'_{V_2}} \\
    \implies \textcolor{device}{O'_D} = \textcolor{device}{C'_D} \, \textcolor{vps2}{C_{V_2}'^{-1} \, O'_{V_2}}
\end{equation}

Now we know the poses of the objects $O$ and $O'$ in the same frame, \textcolor{device}{$D$}. The relative positions of these objects will remain the same in any coordinate frame of reference:

\begin{equation}
     \textcolor{device}{O_D^{-1} \, O'_D} = O_M^{-1} \, O'_M, \; \forall M
\end{equation}

Although the objects $O$ and $O'$ are in different rooms and scanned by different VPS services, we now know the relative pose of one with respect to the other.

\begin{align*}
    O_D^{-1} \, O'_D &= (C_D \, C_{V_1}^{-1} \, O_{V_1})^{-1} \, (C'_D \, C_{V_2}'^{-1} \, O'_{V_2}) \\
    &=O_{V_1}^{-1} \, C_{V_1} \, C_D^{-1} \, C'_D \, C_{V_2}'^{-1} \, O'_{V_2}
\end{align*}

As the locations of objects $O$ and $O'$ in $V_1$ and $V_2$ is arbitrary, we choose $O_{V_1} = O'_{V_2} = I$. In other words, the relative transform between the origins of the two coordinate frames $V_1$ and $V_2$ is: 

\begin{equation}
    \textcolor{vps1}{C_{V_1}} \, \textcolor{device}{C_D^{-1} \, C'_D} \, \textcolor{vps2}{C_{V_2}'^{-1}}
\label{eqn:mapTransformEqn}
\end{equation}

\begin{figure}
    \centering
    \includegraphics[width=0.85\linewidth]{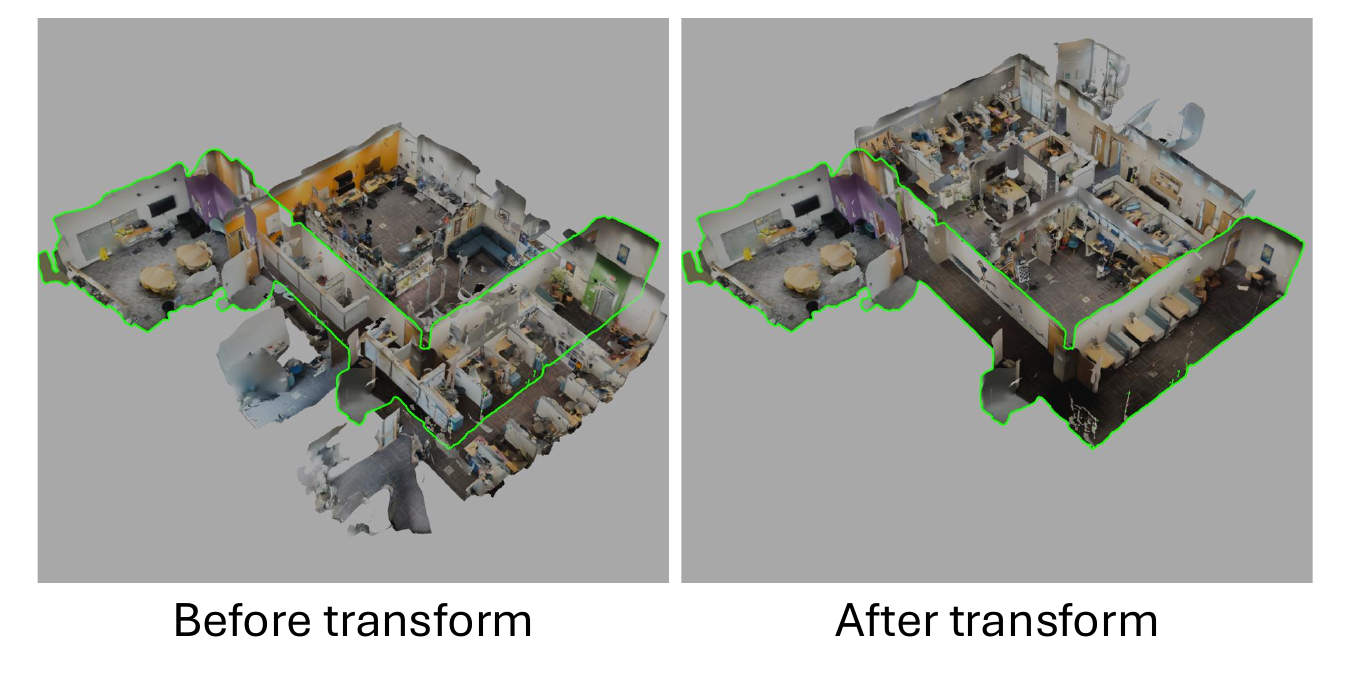}
    \caption{Example of stitching coordinate systems.}
    \label{fig:dynamicStitchingExample}
\end{figure}

The matrices $C_{V_1}$, $C_{V_2}$, and $C_D$ can be repeatedly measured by the device and collected from the VPS services as the device moves between rooms 1 and 2. As the matrices are noisy because of localization errors, it might take a few iterations of measurements to get the right transform. Figure~\ref{fig:dynamicStitchingExample} shows the relative transform (i.e., Transform~\ref{eqn:mapTransformEqn}) between the 3D scans of two independent VPS services. We show a green outline around one of the scans. In \S~\ref{eval:dynStitching}, we show that only one observation from the second room is sufficient, in most cases, to align the two coordinate frames of reference, resulting in seamless transitions between VPS services.

It is important to note that the transformation described in this section applies only to linear transformations between 3D scans and does not account for non-linear warps. However, in our 3D map construction pipeline, we remove distortion from our input images and use the \textit{corrected} images to generate the maps, effectively removing non-linear distortions. 
\section{3D Scanning pipeline}
\label{sec:scanning}

This section outlines the 3D scanning pipeline used for constructing the VPS map in \systemname{}. Our pipeline consists of data acquisition with mobile devices, filtering dynamic elements, and building a feature database for localization. 

\textbf{Collecting posed images:}  
We begin by scanning the target environment using an iPad equipped with a LiDAR sensor, capturing data through the Polycam app. This provides us with RGB-D images alongside 6-DoF camera poses. To ensure geometric accuracy, the iPad’s camera intrinsics are calibrated using Zhang’s method~\cite{zhang2002flexible}. The RGB-D images exported by Polycam are already undistorted to conform to the pinhole camera model, which is essential for accurate 3D to 2D projection.

\textbf{Dynamic object removal:}  
To improve map stability over time, we remove dynamic elements from the scene. We use YOLO-v8~\cite{yolo} for semantic segmentation and classification, generating masks that identify potentially dynamic objects such as people, chairs, and cars. These masks are applied to the RGB images to exclude dynamic regions from the mapping process, ensuring the VPS map captures only the static structure of the scene.

\textbf{Feature database construction:}  
The masked images are then processed using COLMAP~\cite{schonberger2016structure} to generate a 3D reconstruction and image feature database. We replace COLMAP’s default feature pipeline with SuperPoint for keypoint extraction and SuperGlue for matching, which provides more robust performance. Additionally, we compute a NetVLAD global descriptor for each image to accelerate image retrieval during localization.

\section{Implementation and support tools}

\begin{figure}
\centering
    \begin{subfigure}{.2\textwidth}
        \centering
        \includegraphics[width=.95\linewidth]{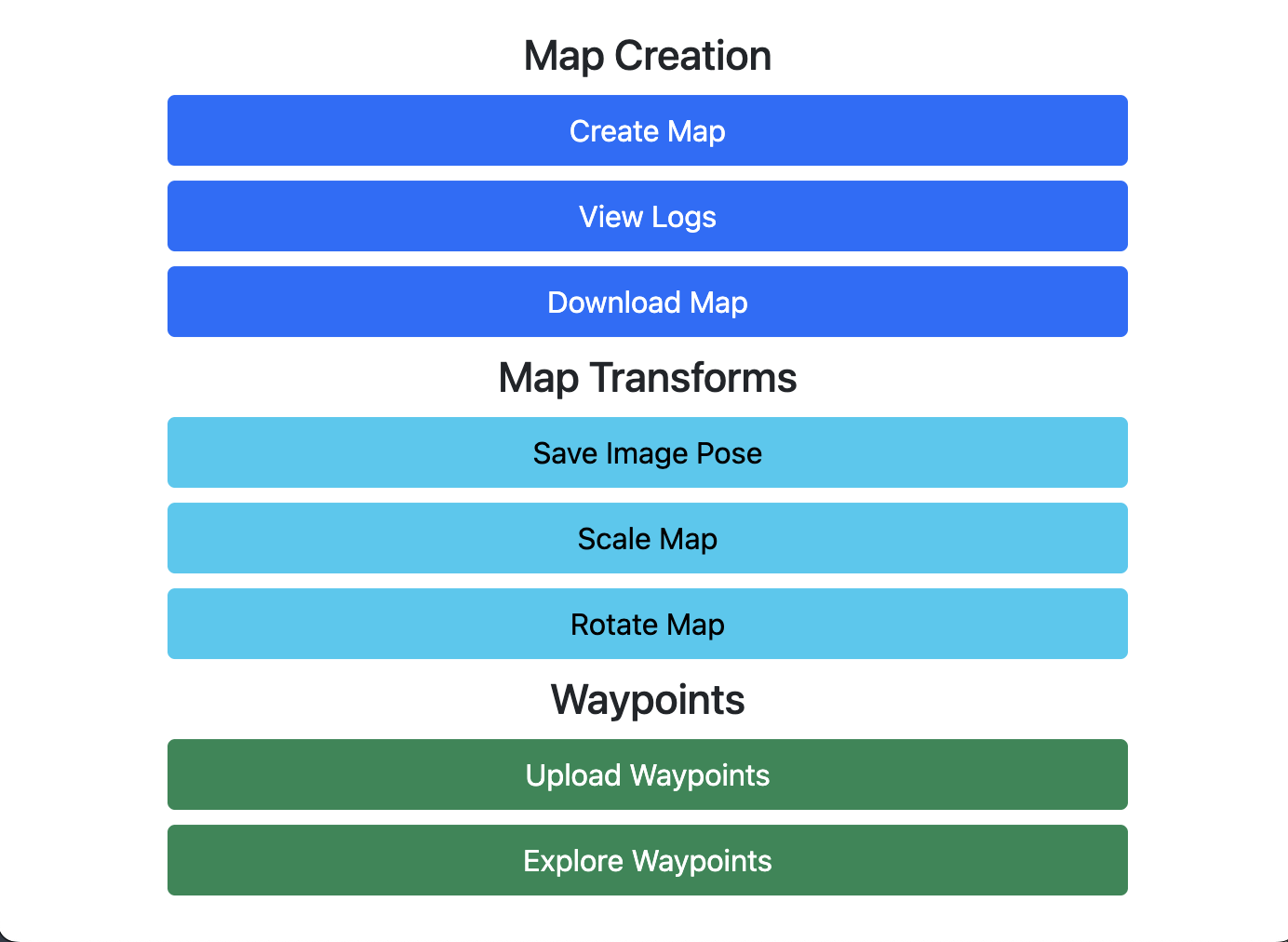}  
        \caption{VPS Service landing page.}
        \label{fig:vpsServiceLandingPage}
    \end{subfigure}
    \hfill
    \begin{subfigure}{.2\textwidth}
        \centering
        \includegraphics[width=.95\linewidth]{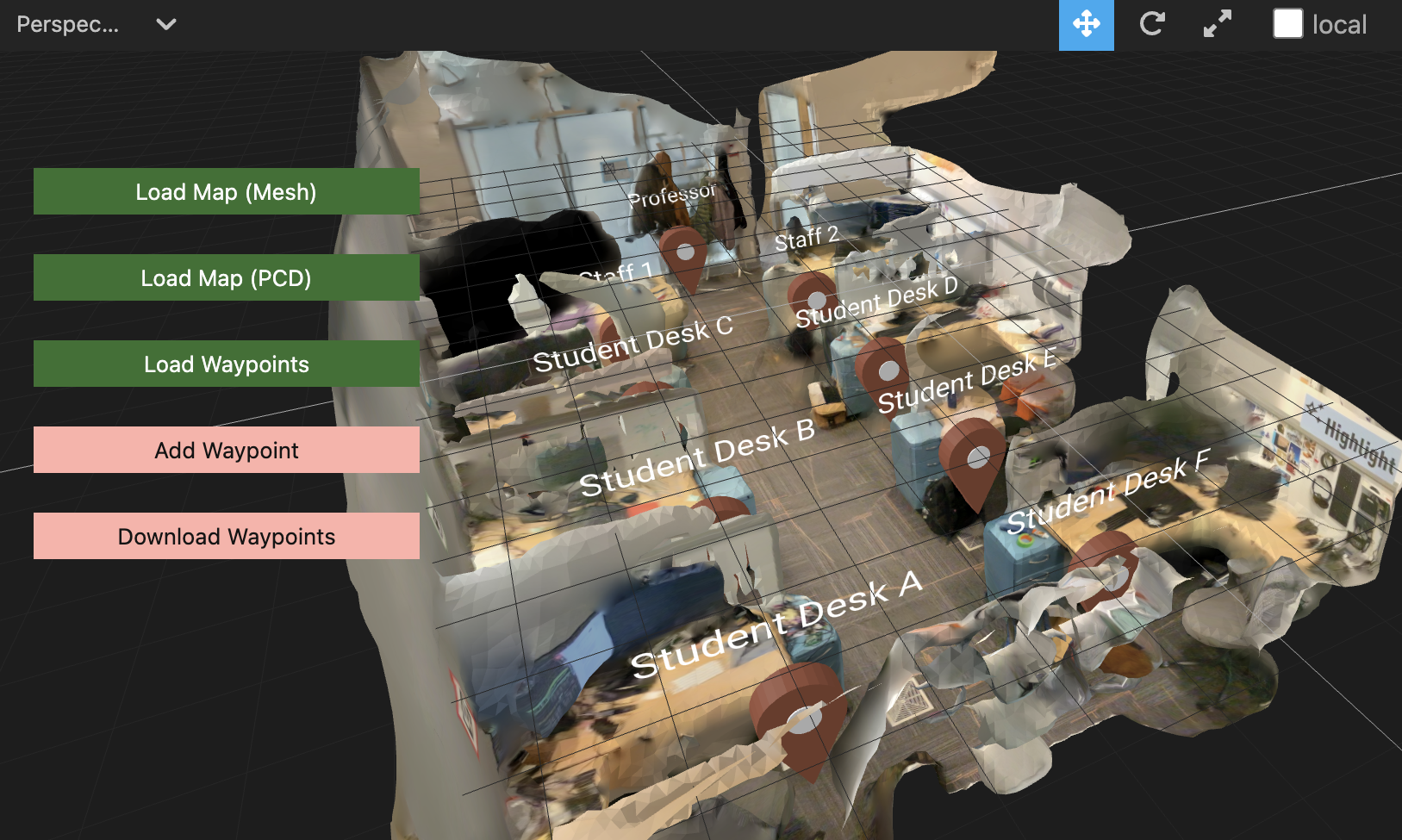}  
        \caption{Scan Visualizer and tagging.}
        \label{fig:scanVisualizer}
    \end{subfigure}
    \hfill
    \begin{subfigure}{.2\textwidth}
        \centering
        \includegraphics[width=.95\linewidth]{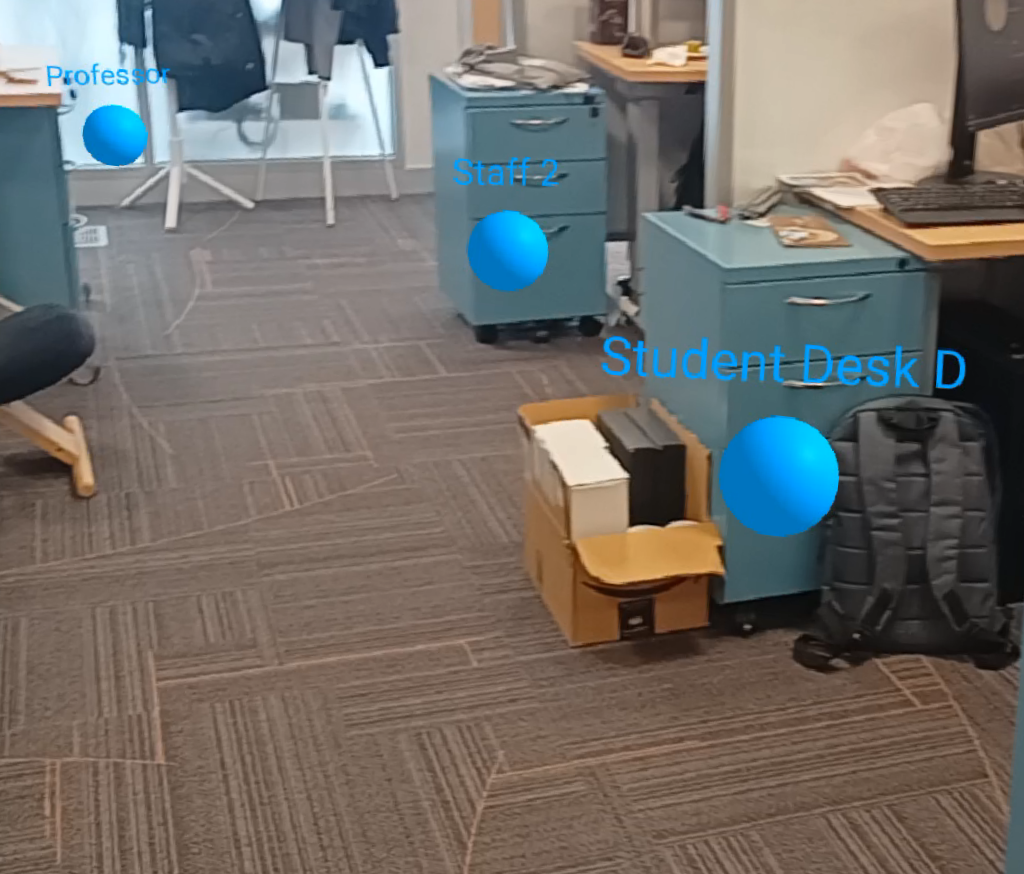}  
        \caption{AR localization verifier.}
        \label{fig:arVerifier}
    \end{subfigure}
    \hfill
    \begin{subfigure}{.2\textwidth}
        \centering
        \includegraphics[width=.95\linewidth]{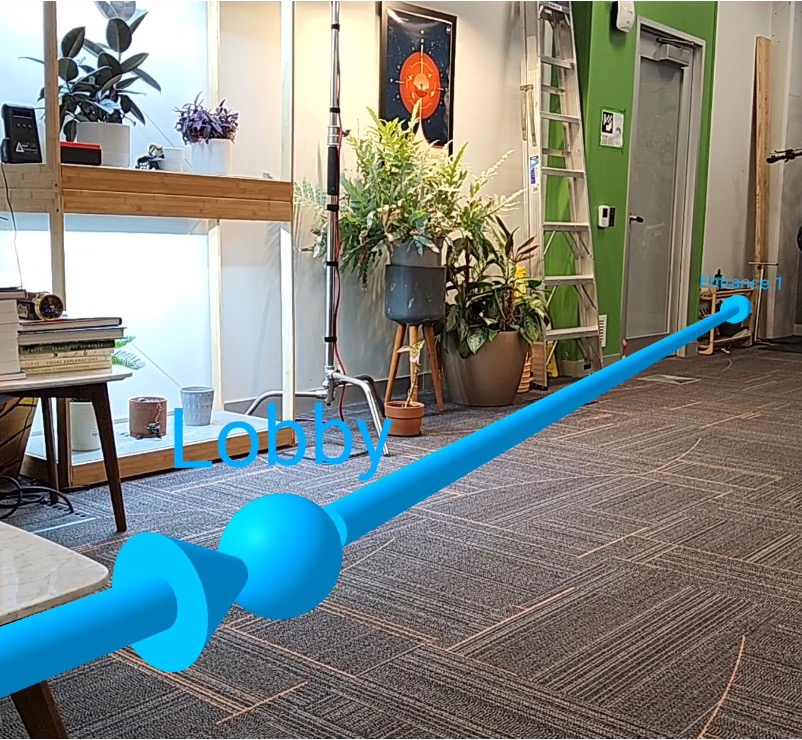}  
        \caption{AR Navigation application.}
        \label{fig:navigationApp}
    \end{subfigure}
\caption{Supplementary tools.}
\label{fig:tools}
\end{figure}

We will open-source \systemname{}---both device-side and VPS service implementations. We also present supplementary tools that help with 3D scan visualization and verification of pose estimation. Additionally, we have implemented a 3D indoor navigation application that uses \systemname{} as the underlying localization backend to show that it can support large-scale AR applications.

\textbf{VPS service} (\ref{fig:vpsServiceLandingPage}): Our implementation of VPS service uses an optimized version of hloc~\cite{sarlin2019coarse} for pose estimation (\S~\ref{sec:PoseEstimation}), along with CLIP and YOLO models for other components described in \S~\ref{sec:lolPipeline}. In our evaluations, it runs on an Intel Core i9-13900K CPU and an NVIDIA GeForce RTX 4090 GPU.

\textbf{Scan visualization and tagging} (\ref{fig:scanVisualizer}): We present a web-based tool built using the A-Frame Inspector~\cite{a_frame_inspector} that can visualize 3D scans generated for the VPS service (\S~\ref{sec:scanning}). It also allows tagging regions in these scans. These tags can be downloaded and rendered against the real-world using our AR verifier tool below.

\textbf{AR localization verifier} (\ref{fig:arVerifier}): An A-Frame~\cite{a_frame} application that uses camera RGB images and the selected VPS service to localize. It overlays the tags created using the above tool onto the real world. 
It enables quick verification that the VPS service can correctly anchor virtual objects to the physical environment. Figure~\ref{fig:arVerifier} shows the tags in \ref{fig:scanVisualizer} overlaid on the real-world.

\textbf{AR 3D indoor navigation application}(\ref{fig:navigationApp}): A-Frame application that uses \systemname{} as the localization backend to navigate large indoor spaces. Between remote localization calls to \systemname{}, it uses WebXR for local VIO tracking. The data for route calculation is generated using the scan visualizer tool above. Details on routing and navigation in our application is out of scope of this paper.
\section{Evaluation}
\label{sec:evaluation}

In this section, we evaluate our techniques to demonstrate the feasibility of a federated VPS system. While each method can be further optimized in future work, we believe our results show that a federated VPS solution is viable and can support the needs of future augmented reality applications.

\textbf{Dataset}: To evaluate \systemname{}, we 3D-scanned and set up 30 VPS services for various indoor locations on a university campus. The locations are diverse, including conference rooms, office cubicles, building lobbies, and classrooms. We used the raw data export from the Polycam application on an iPad Pro to collect our scans. We avoid using expensive scanning equipment to demonstrate that our techniques are effective, even with low-quality, easily obtainable scans. Each indoor location contains many movable objects, such as chairs, tables, keyboards, backpacks, etc., that change position over time. For each location, we collect two sets of query images---one immediately after the scan, and another after moving the objects in the scene. We use these query sets to show that our technique works for dynamic spaces without imposing the need to re-scan them. Each query image has an AprilTag~\cite{wang2016apriltag} to get baseline pose estimates. To evaluate our techniques that rely on local VIO traces, we record WebXR poses in our AR application as a test device moves through spaces.


\subsection{Visual features-free dynamic VPS stitcher}
\label{eval:dynStitching}

\begin{figure}[htp]
    \centering
    \begin{subfigure}{.23\textwidth}
        \centering
        \includegraphics[width=1.0\linewidth]{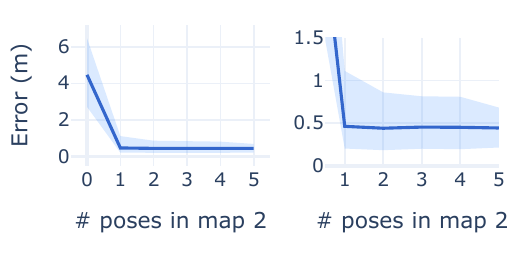}  
        \caption{Translation Error.}
        \label{fig:eval:transDynStitching}
    \end{subfigure}
    \hfill
    \begin{subfigure}{.23\textwidth}
        \centering
        \includegraphics[width=1.0\linewidth]{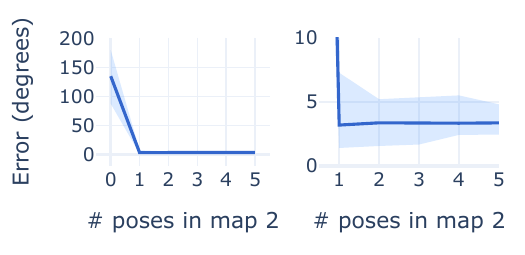}  
        \caption{Rotation Error.}
        \label{fig:eval:rotDynStitching}
    \end{subfigure}
    \caption{\ifhl{Error in the stitching transform estimated by the VPS Stitcher component compared to the \textit{true stitching transform}---transform calculated by manually stitching the two maps.}}
    \label{fig:dynStitchingErrors}
\end{figure}

In \S~\ref{sec:dynStitching}, we described our technique for dynamically stitching coordinate systems from independent VPS services, without requiring the VPS services to share visual features with each other. This technique requires pose matrices from three sources---the two VPS services being stitched and the local VIO tracking results.

Ideally, the device should determine the correct transform between two VPS systems as soon as it transitions from one service to the other, without needing numerous localizations in the new service. We use the VIO traces and localization results on images in our query dataset, and apply Transform Expression~\ref{eqn:mapTransformEqn} on them, and compare the result with the true transform---the transform we get by manually aligning independent VPS maps.  Figure~\ref{fig:dynStitchingErrors} shows the translation and rotation errors between the true transform and the transform calculated using (\ref{eqn:mapTransformEqn}) against the number of pose estimations in the second VPS map. The shaded area shows $5^{th}$ and $95^{th}$ percentiles. We see that the device fixes on a good transform just after a single query to the second VPS service. The translation error on average goes down from more than 4~m to 0.5~m, and the rotation error (i.e., angle-axis distance: $cos^{-1} \frac{\text{trace}{R_1R_2} - 1}{2}$, where $R_1$ and $R_2$ are 3X3 rotation matrices) goes down from 180\textdegree{} in the worst case to about 5\textdegree{} immediately after the first localization result. In this evaluation, we consider the 5 best poses from the first VPS service when calculating the transform in (\ref{eqn:mapTransformEqn}). Concretely, we use 5 values of $C_{V_1}$ and $C_D$, resulting in 5 different transformations which are then averaged to get the required transform. While considering more localization results after entering the second VPS service does not significantly reduce the median errors, the worst-case error (i.e., $95^{th}$ percentile in the figures) decreases.

\subsection{Masked localization in dynamic environments}
\label{sec:eval:maskedLocalization}

\begin{figure}
\centering
    \begin{subfigure}{.23\textwidth}
        \centering
        \includegraphics[width=1.0\linewidth]{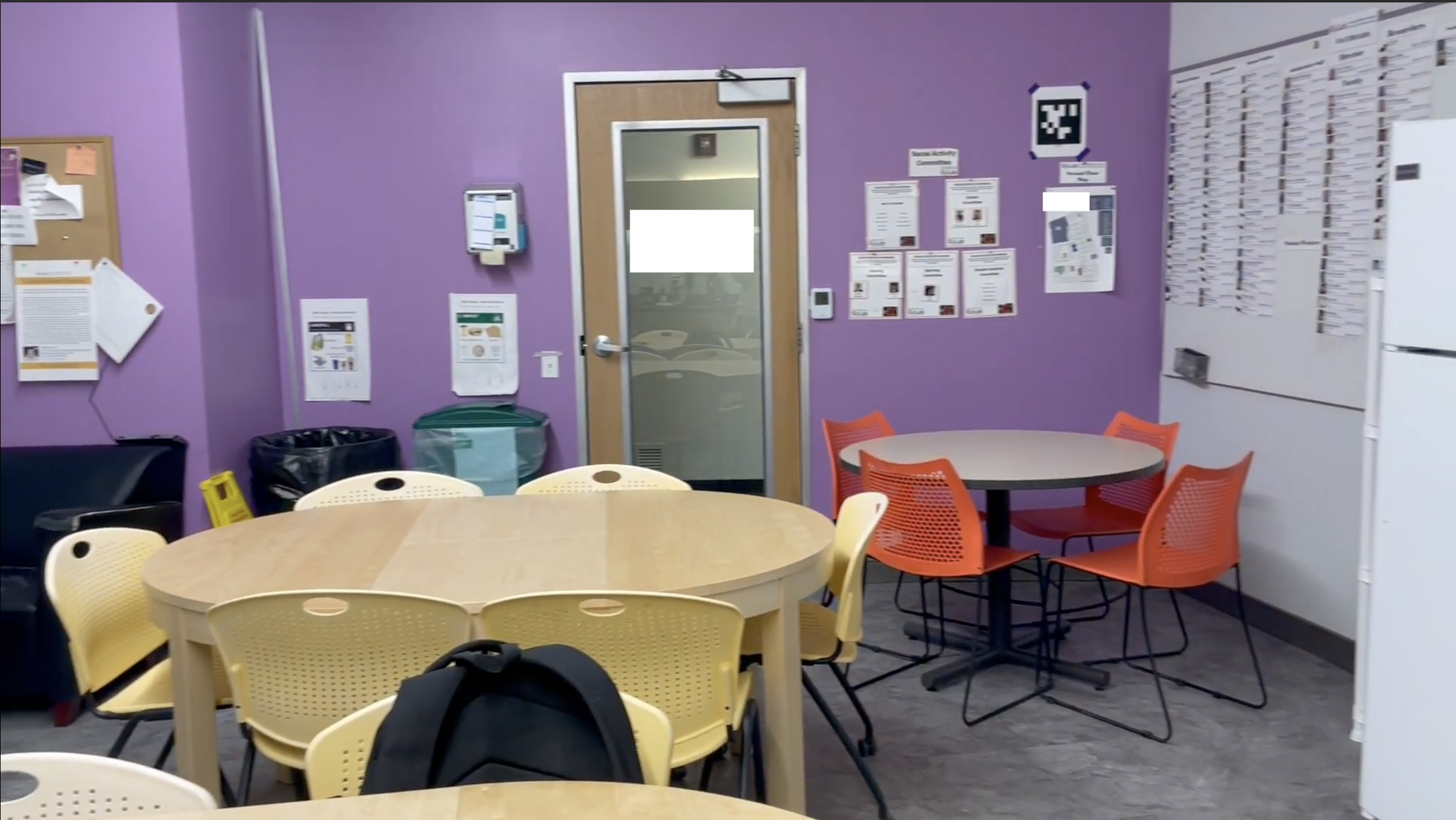}
        \caption{Image used for map creation.}
        \label{fig:mapFrame}
    \end{subfigure}
    \hfill
    \begin{subfigure}{.23\textwidth}
        \centering
        \includegraphics[width=1.0\linewidth]{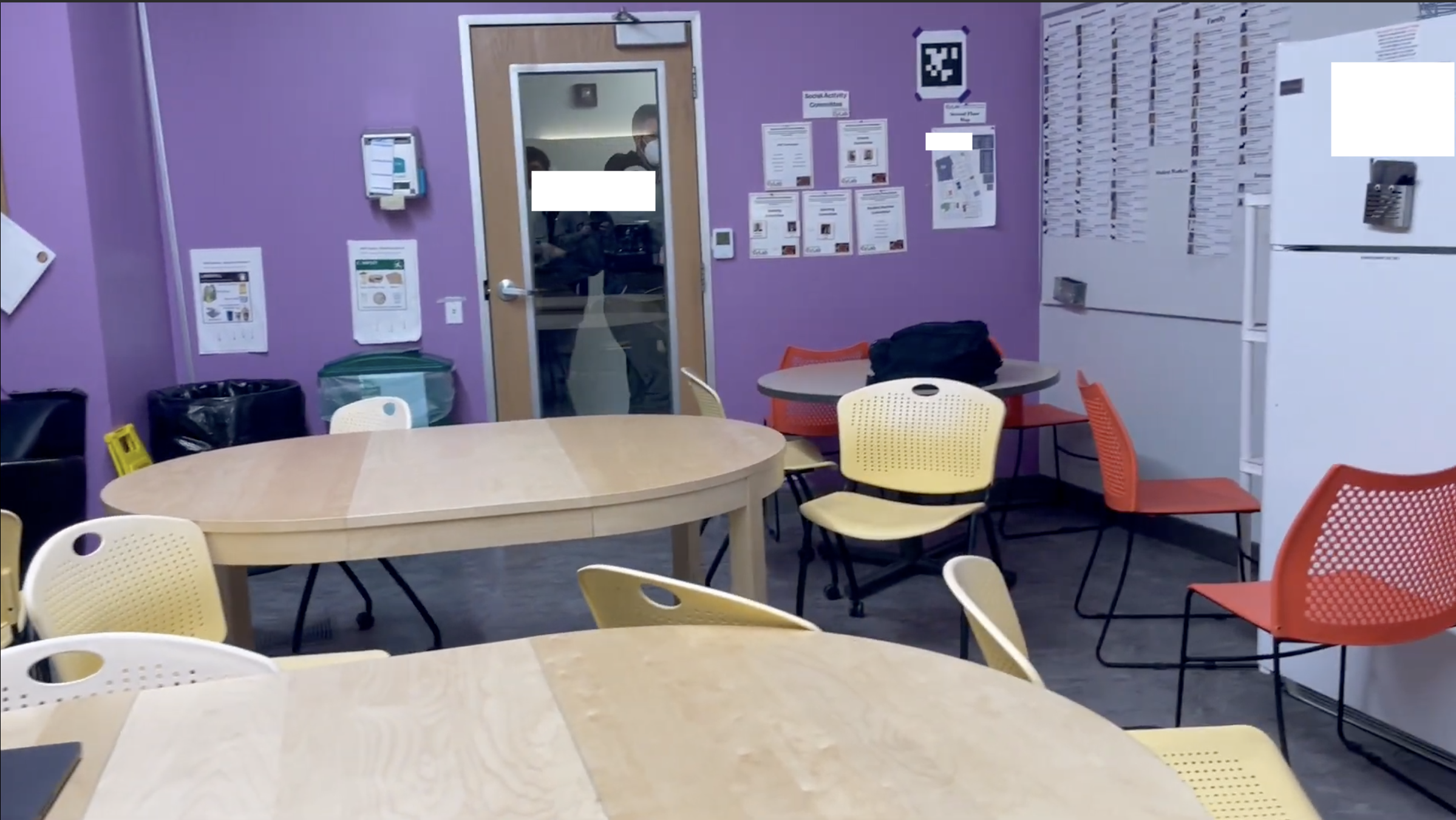}
        \caption{Query image.}
        \label{fig:queryFrame}
    \end{subfigure}
\caption{Differences between the query image and the image used for map creation to evaluate our masked localization pipeline.}
\label{fig:queryMapFrames}
\end{figure}

To evaluate our masked localization pipeline, we compare poses estimated by our pipeline against ground-truth poses obtained using AprilTags. The evaluation uses query images captured in environments where objects such as chairs and backpacks were moved since the map was created. Figure~\ref{fig:queryMapFrames} shows an example comparison between a query image (\ref{fig:queryFrame}) and the corresponding map image (\ref{fig:mapFrame}). Notice that the chairs and backpacks are in different positions in the two images. Such queries allow us to evaluate how our masked localization pipeline handles dynamism.

\begin{figure}
\centering
    \begin{subfigure}{.23\textwidth}
        \centering
        \includegraphics[width=0.95\linewidth]{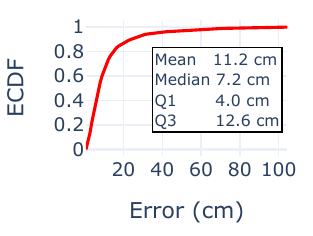}
        \caption{Translation RPE}
        \label{fig:rpeTranslation}
    \end{subfigure}
    \hfill
    \begin{subfigure}{.23\textwidth}
        \centering
        \includegraphics[width=0.95\linewidth]{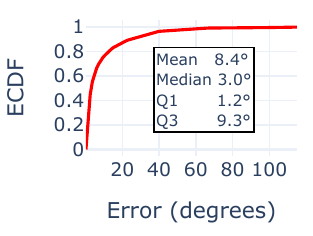}
        \caption{Rotation RPE}
        \label{fig:rpeRotation}
    \end{subfigure}
\caption{Relative Pose Error (w.r.t. AprilTags).}
\label{fig:rpeAprilTag}
\end{figure}

Each query image contains an AprilTag, allowing us to compute two poses: one using the AprilTag and another using our localization pipeline. We generate two trajectories per query set, one from AprilTag poses and one from our method, and align them before computing the relative pose error (RPE). Figure~\ref{fig:rpeAprilTag} shows the empirical cumulative distribution function (ECDF) of the translation and rotation RPEs compared to AprilTag poses. The median translation error is 7.2~cm, and the median rotation error is 3\textdegree, which is acceptable for most AR applications.

\begin{figure}[t]
\centering
    \begin{subfigure}{.2\textwidth}
        \centering
        \includegraphics[width=0.95\linewidth]{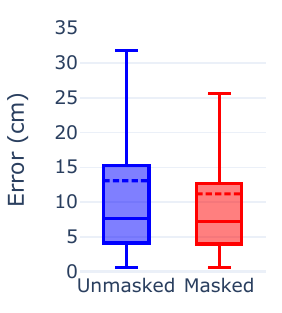}
        \caption{Translation errors.}
        \label{fig:maskedVsUnmaskedTrans}
    \end{subfigure}
    \hfill
    \begin{subfigure}{.2\textwidth}
        \centering
        \includegraphics[width=0.95\linewidth]{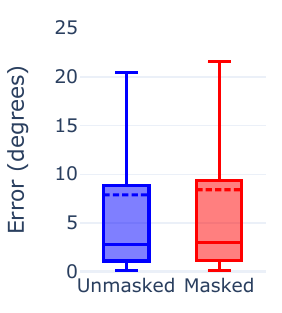}
        \caption{Rotation errors.}
        \label{fig:maskedVsUnmaskedRot}
    \end{subfigure}
\caption{Performance of masked localization.}
\label{fig:maskedVsUnmasked}
\end{figure}

Figure~\ref{fig:maskedVsUnmasked} compares errors observed with and without masking. The box plots (whiskers extend to 1.5 times the interquartile range beyond the first and third quartiles) shows the distribution of errors. Masking reduces translation errors, especially in worst-case scenarios---the 95\textsuperscript{th} percentile error is reduced by 21.1\% and the mean error decreases by 14.4\%. In case of rotation, we see a slight increase in worst case errors---the 95\textsuperscript{th} percentile error increases by 2.5\%, while the mean stays about the same.

\subsection{Pose confidence}
\label{sec:eval:poseConf}

\begin{figure}
    \centering
    \includegraphics[width=0.95\linewidth]{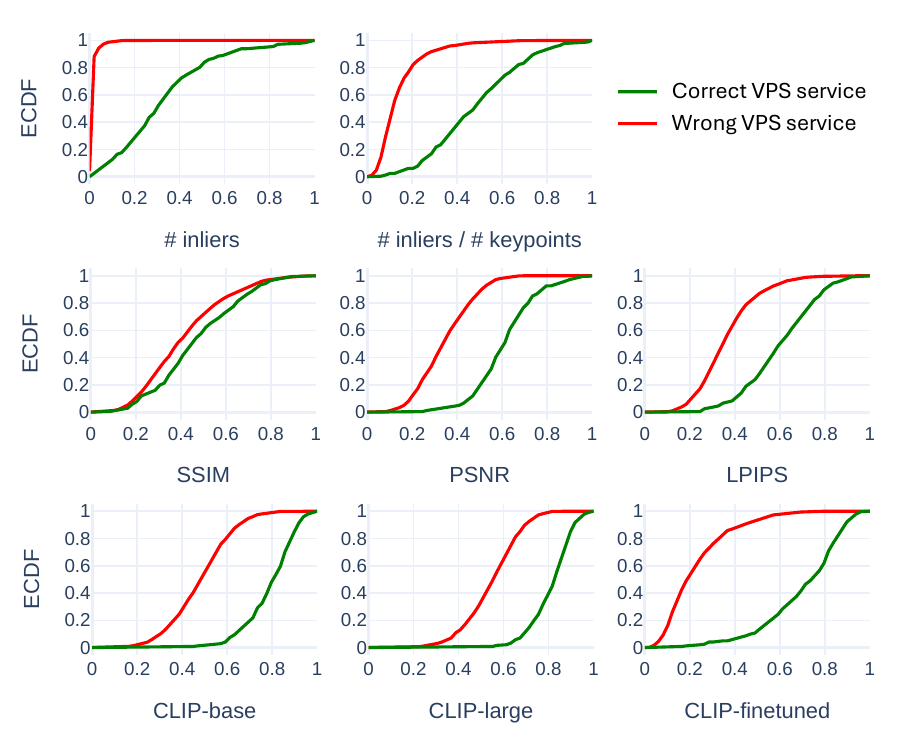}
    \caption{CDFs of metrics estimating pose confidences. \ifhl{The CLIP-finetuned model shows the best discriminability between correct and wrong VPS services.}}
    \label{fig:poseConfMetricCDFs}
\end{figure}

As described in \S~\ref{sec:poseConfidence}, we require a pose confidence metric that is comparable across VPS services, agnostic to individual service configurations, and capable of distinguishing correctly localized images from potentially inaccurate ones.

In this section, we compare several confidence metrics and justify our choice of a fine-tuned CLIP model. Figure~\ref{fig:poseConfMetricCDFs} shows the cumulative distribution functions (CDFs) of confidence scores for various metrics we evaluated. The green line represents scores when queries are made to the correct VPS service, while the red line corresponds to queries sent to an incorrect service. All metrics are normalized to the range [0, 1], and we invert LPIPS so that higher values indicate greater confidence. A good confidence metric must easily distinguish between correct and wrong VPS services. In other words, good metrics have large gaps between the green and the red line in Figure~\ref{fig:poseConfMetricCDFs}.

The first two rows show feature-based metrics: the number of SuperPoint inliers and the ratio of inliers to the total number of detected SuperPoint keypoints. These metrics are tightly coupled to the specific feature detector (SuperPoint) and matcher (SuperGlue) used in our pose estimation pipeline and are not directly comparable across VPS services; they are shown here for reference. We observe that, within the same pose estimation setup, these feature-based metrics provide good discriminability between correct and incorrect matches. The second row presents image-based comparison metrics between the rendered and query images, as described in \S~\ref{sec:poseConfidence}. Metrics such as SSIM, PSNR, and LPIPS show limited ability to distinguish between true and false matches when compared to feature-based approaches. The third row shows the cosine similarity of CLIP image encodings. ‘CLIP-finetuned’ refers to our version of CLIP, fine-tuned on our dataset to produce similar embeddings for corresponding rendered and query images. Among all metrics, the fine-tuned CLIP model demonstrates the best discriminability.

\begin{figure}
    \centering
    \includegraphics[width=0.8\linewidth]{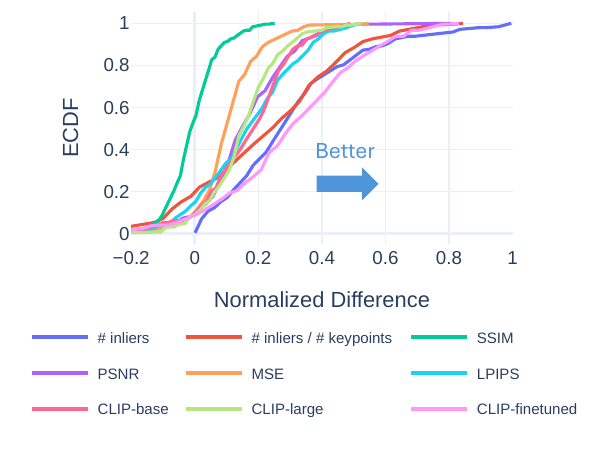}
    \caption{Discriminability of pose confidence metrics.}
    \label{fig:poseConfDiscriminability}
\end{figure}

Figure~\ref{fig:poseConfDiscriminability} presents the empirical cumulative distribution functions (ECDFs) of normalized confidence score differences between correct and incorrect VPS queries across various pose confidence metrics. A higher curve indicates better discriminability between true and false localizations, as more samples exhibit a larger difference in confidence scores. Notably, the CLIP-finetuned method demonstrates the highest discriminative power supporting our decision to use it as our pose confidence metric.

\subsection{VPS Selector}
\label{sec:eval:vpsSelector}

In \S~\ref{sec:VPSSelector}, we described how the device compares a remote trajectory from a VPS service with its own local VIO trajectory to select the appropriate VPS service. In this section, we show that this approach enables correct VPS selection after only a few localization cycles.

\begin{figure}[htp]
    \centering
    \begin{subfigure}{.23\textwidth}
        \centering
        \includegraphics[width=1.0\linewidth]{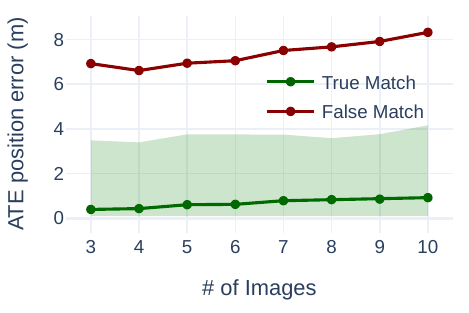}
        \caption{ATE between local VIO trajectory and remote service trajectory.}
        \label{fig:VPSSelectorATETraj}
    \end{subfigure}
    \begin{subfigure}{.23\textwidth}
        \centering
        \includegraphics[width=1.0\linewidth]{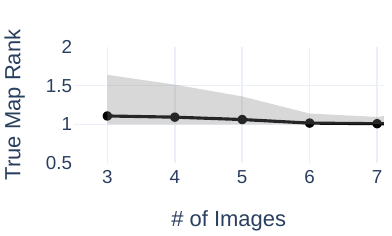}
        \caption{Mean rank of the correct VPS service.}
        \label{fig:VPSSelectorRank}
    \end{subfigure}
    \caption{Performance of VPS Selector. \ifhl{The correct map is reliably ranked at the top even with few localization cycles.}}
    \label{fig:eval:VPSSelector}
\end{figure}

In this evaluation, we group five VPS services located in close proximity—such that they are all discovered by a single discovery query—and assess whether our VPS selector can correctly identify the appropriate service using only local VIO traces. Figure~\ref{fig:VPSSelectorATETraj} shows the absolute trajectory error (ATE) in position between the local VIO trajectory and the VPS-estimated trajectory, for both correct and incorrect service selections. The green shaded region represents the spread between the 5th and 95th percentile errors for correct matches. We observe that the ATE for incorrect matches increases as more images are localized. The 95th percentile error for true matches is lower than the median error for false matches, showing our technique's reliability in selecting the right candidate.


Figure~\ref{fig:VPSSelectorRank} illustrates the mean rank of the correct VPS service against the number of localization cycles. A rank of one indicates that the service was selected as the top candidate by \systemname{} on the device. Even with a small number of localization cycles, the correct VPS service consistently achieves a mean rank close to one. As the length of the compared trajectory increases, the variability in the estimated rank decreases, indicating more reliable selection.

\subsection{Place recognizer}
\label{sec:eval:placeRecognizer}

To evaluate the Place Recognizer module (\S~\ref{sec:placeRecognizer}), we consider a group of five VPS services covering places that are close and look visually similar to one another. Figure~\ref{fig:placeRecognizer} shows the confidence scores produced by the Place Recognizer module for various rooms. For each query, a CLIP model fine-tuned to each individual room is used to compute a confidence score, and the results are visualized as box plots across six rooms. The green box in each subplot represents the correct room for the query, while the red boxes indicate other (incorrect) rooms.

\begin{figure}
    \centering
    \includegraphics[width=0.8\linewidth]{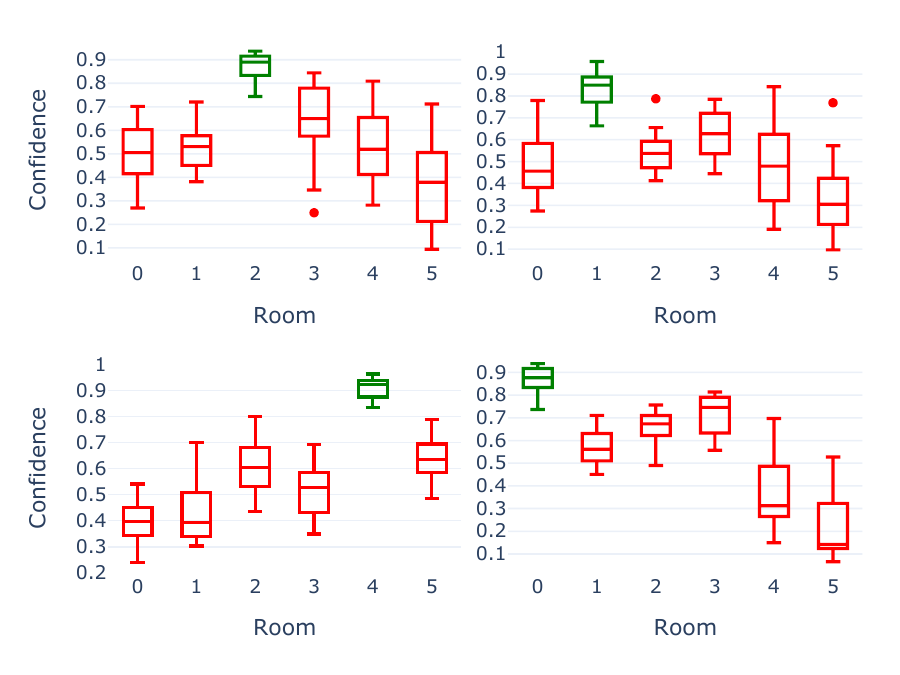}
    \caption{Performance of Place Recognizer module.}
    \label{fig:placeRecognizer}
\end{figure}

Across all four subplots, the correct room consistently achieves the highest confidence score, clearly separated from the others. This demonstrates the discriminative power of the fine-tuned CLIP models in correctly identifying the room associated with a given query. The spread of the red boxes also highlights that incorrect rooms receive significantly lower and more varied confidence scores, reinforcing the reliability of the module in distinguishing between similar indoor environments.

\section{Limitations and Future Directions}

The infrastructure presented in this paper is only a step towards our vision of a widely deployed federated VPS infrastructure. Several important challenges require further careful consideration. Two that we consider briefly below are privacy and application design. 

\ifhl{
\textbf{Privacy considerations}: As discussed in \S~{\ref{sec:VPSSelector}}, the VPS Selector module requires poses for at least 3 images to identify the right VPS service for its current location. Therefore, initially, the client device has to broadcast the images of its current location to VPS Services that might not have coverage over the location. This opens the possibility for snooping attacks where a VPS Service might register itself adjacent to a sensitive private space (e.g., a secret lab), collect images sent to it during the discovery phase, and reconstruct the map of the private space. This is a violation of privacy. Past work has proven that it is possible to extract information from just extracted features without raw images~{\mbox{\cite{weinzaepfel2011reconstructing, dosovitskiy2016inverting}}}.

To overcome this challenge, VPS Services hosted for sensitive spaces can can adopt existing work on \textit{privacy preserving localization}~{\mbox{\cite{speciale2019privacy, Moon_2024_CVPR, shibuya2020privacy}}}. They perform localization on \textit{secure features} (e.g., \textit{feature lines}~{\cite{speciale2019privacy}}) sent by the client instead of raw images. These secure features reveal little or no intelligible visual or location information to any party except an authorized server that possesses the correct map of the space. Therefore, even if the client broadcasts these secure features to multiple VPS services, only the service that has the correct map of the private space can decode these features and perform localization. Once the Selector has determined the correct VPS service for the location, the client no longer needs to broadcast visual data to multiple services and can continue interfacing with a single VPS service. In this phase, the VPS service can switch to using raw images so that it can take advantage of features such as semantic image masking running on the VPS service. Integrating existing work on privacy preserving localization with \systemname{} is one of our future goals.
}

\textbf{Application Design}: Applications that use one of the existing ubiquitous VPS providers can anchor their content using latitudes, longitudes, and altitudes as their scans are laid out in the geographic coordinate system. As \systemname{} does not have a unified global coordinate system, designing applications on top of \systemname{} is tricky. A straight-forward solution is for applications to position their content using local coordinates of a VPS service and store a pointer to the corresponding VPS service. However, this makes the content tightly coupled with the VPS service, making it unusable when the VPS service is upgraded or replaced. Avoiding such tight-coupling in application design is an interesting future direction. An example solution would be for the VPS services to expose an additional interface that informs applications of the location of some \textit{landmarks} with respect to their local coordinate system. Hallways, doors, elevators and stairs are examples of landmarks in indoor spaces. The application can then author and store content using landmarks as references. As landmarks do not change when VPS services are upgraded or replaced, content will no longer be coupled with specific VPS services.



\section{Conclusion}

We present \systemname{}, a federated VPS system that allows independent organizations to support VPS for their private spaces. Federation of VPS introduces several challenges, such as selection of the right VPS service at a given location, coherence of localization results across service boundaries, and handling dynamic indoor spaces. In this paper, we provide some effective solutions to these challenges and show that the resulting system can provide efficient and accurate localization. Federation of VPS services paves the path for a ubiquitous localization backend, which will enable world-scale augmented reality applications of the future.

\clearpage


\bibliographystyle{abbrv-doi}

\bibliography{template}

\clearpage

\appendix
\section*{Appendix}
\section{Data and Models Description}

We used many existing pre-trained neural network models in \systemname{}. Some of these models were fine-tuned on our own data for some of our components. We also collected our own data and 3D scans to evaluate our pipeline. In this section, we summarize the models and describe the data used in our evaluation.

\subsection{Map 3D scans}

\begin{figure}[h]
\centering
    \begin{subfigure}{.23\textwidth}
        \centering
        \includegraphics[width=1.0\linewidth]{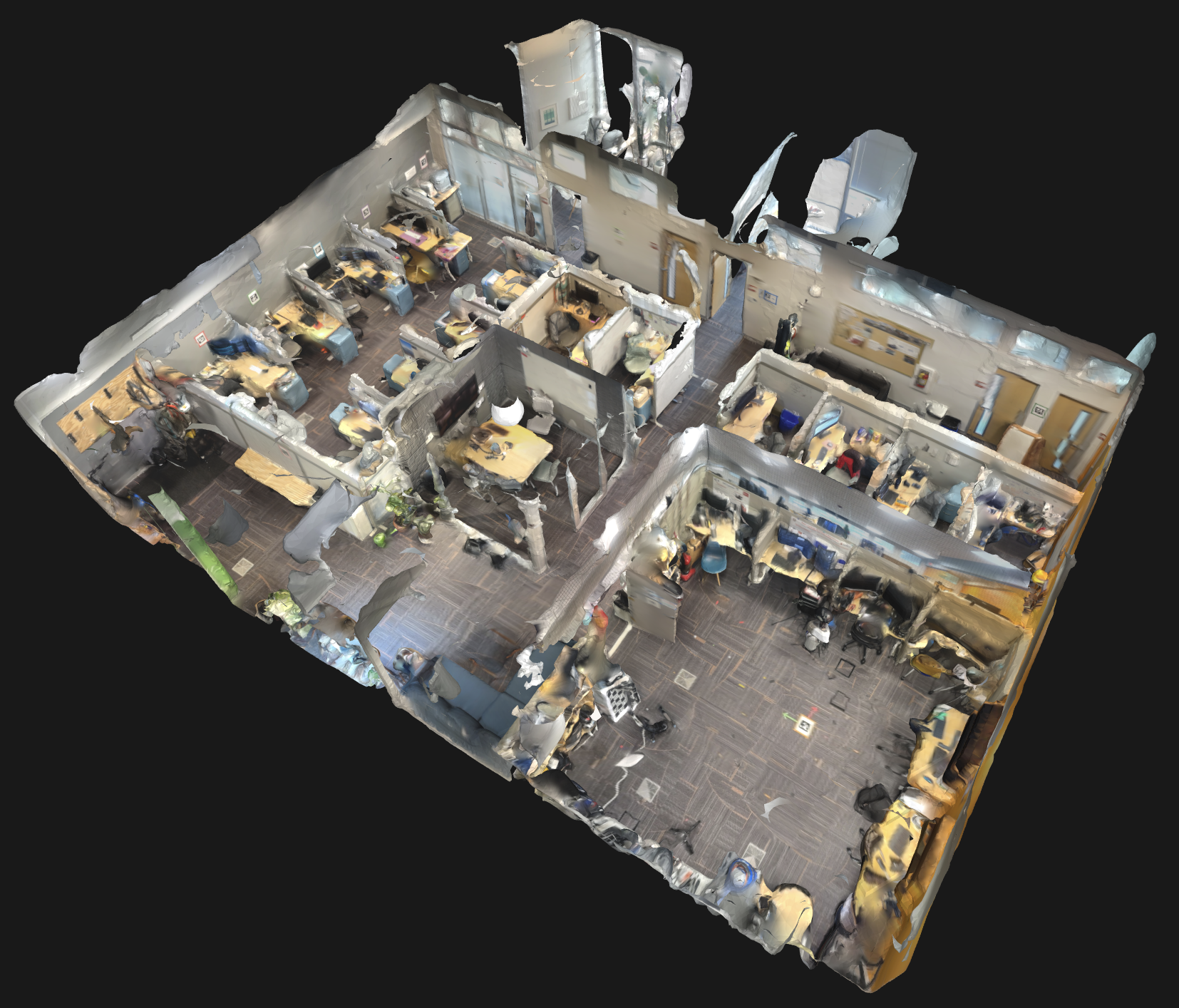}
        \caption{Work Cubicles.}
        \label{fig:cubiclesMap}
    \end{subfigure}
    \hfill
    \begin{subfigure}{.23\textwidth}
        \centering
        \includegraphics[width=1.0\linewidth]{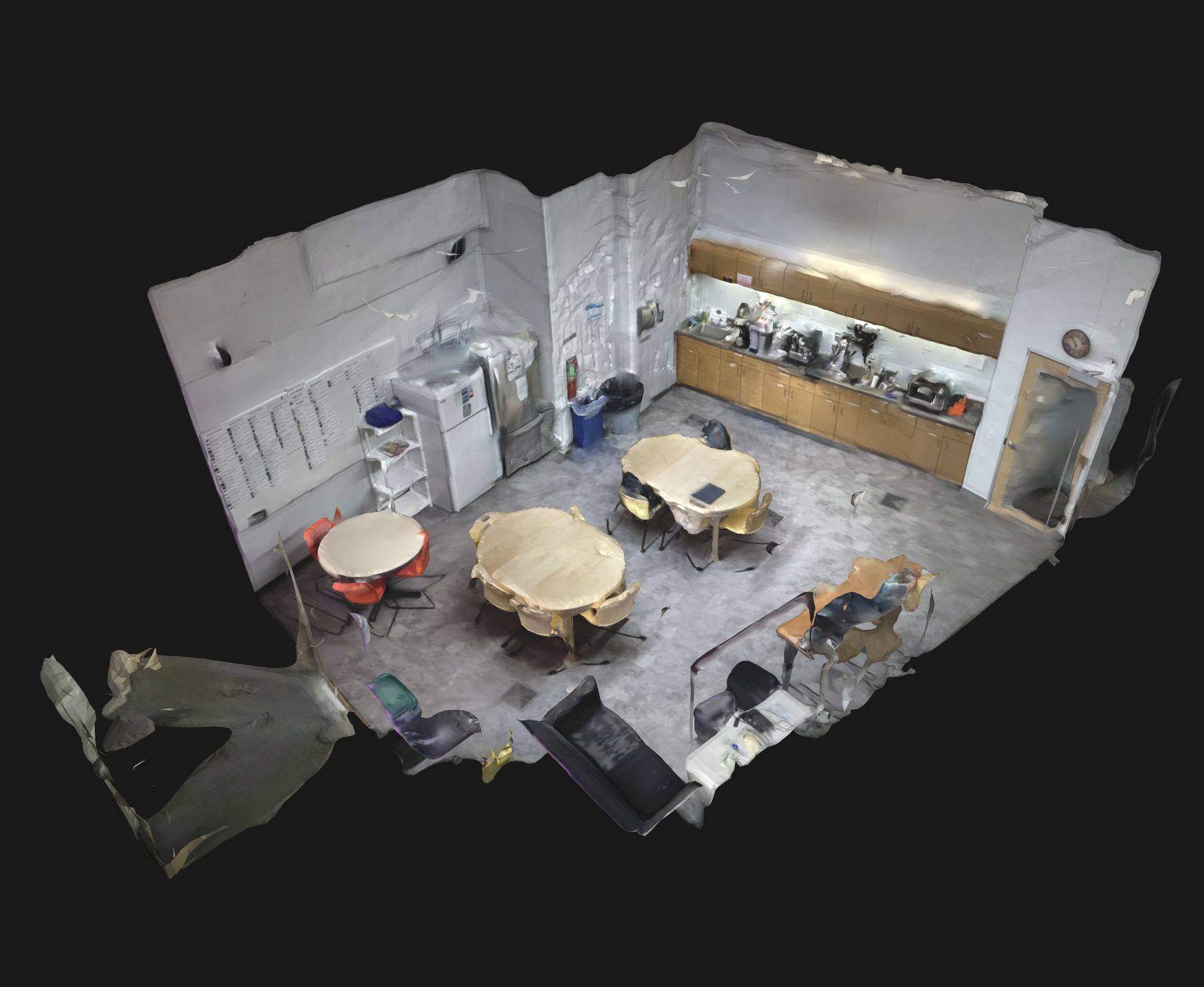}
        \caption{Common Area Kitchen.}
        \label{fig:kitchenMap}
    \end{subfigure}
    \begin{subfigure}{.23\textwidth}
        \centering
        \includegraphics[width=1.0\linewidth]{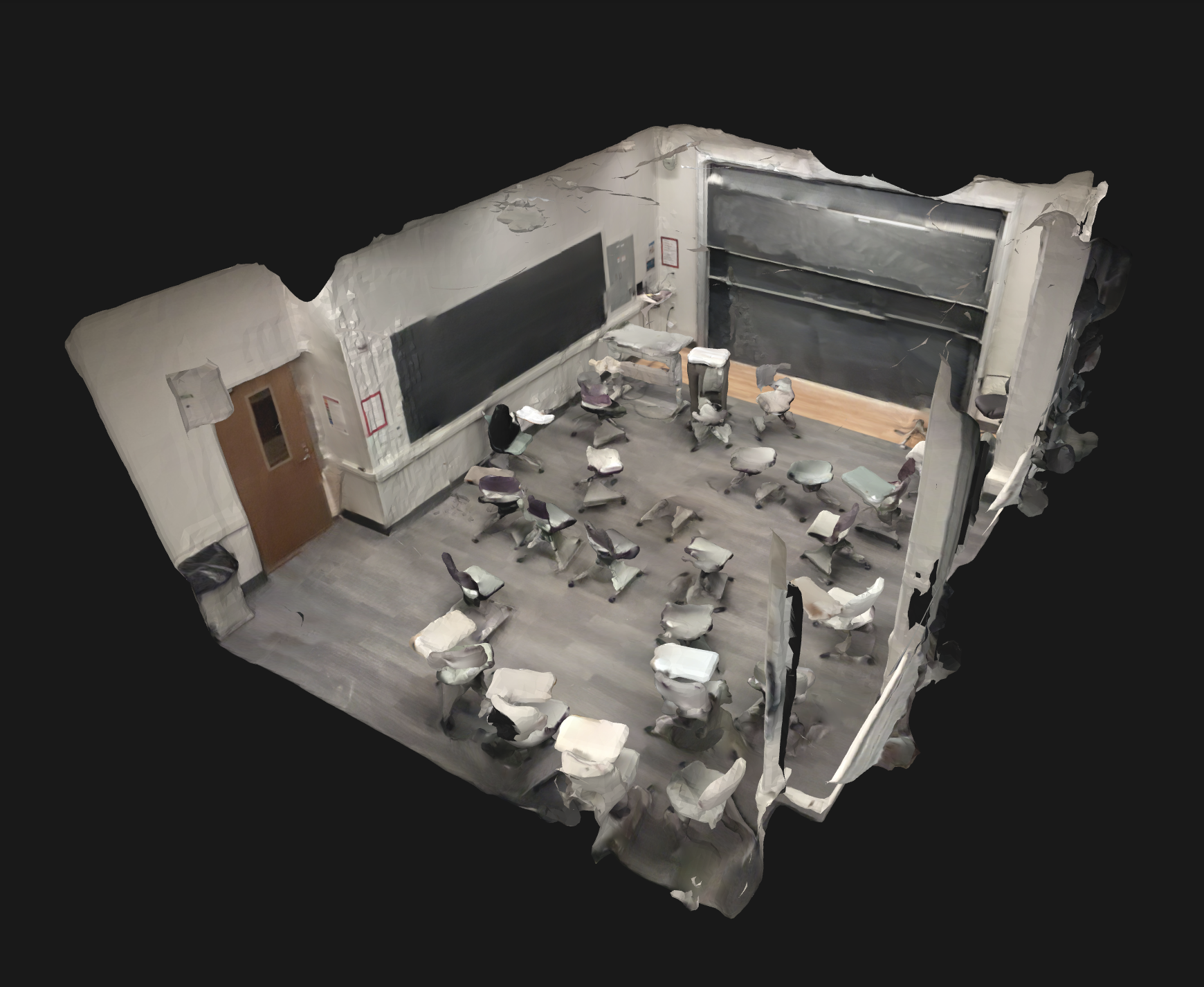}
        \caption{Classroom.}
        \label{fig:classroom}
    \end{subfigure}
    \hfill
    \begin{subfigure}{.23\textwidth}
        \centering
        \includegraphics[width=1.0\linewidth]{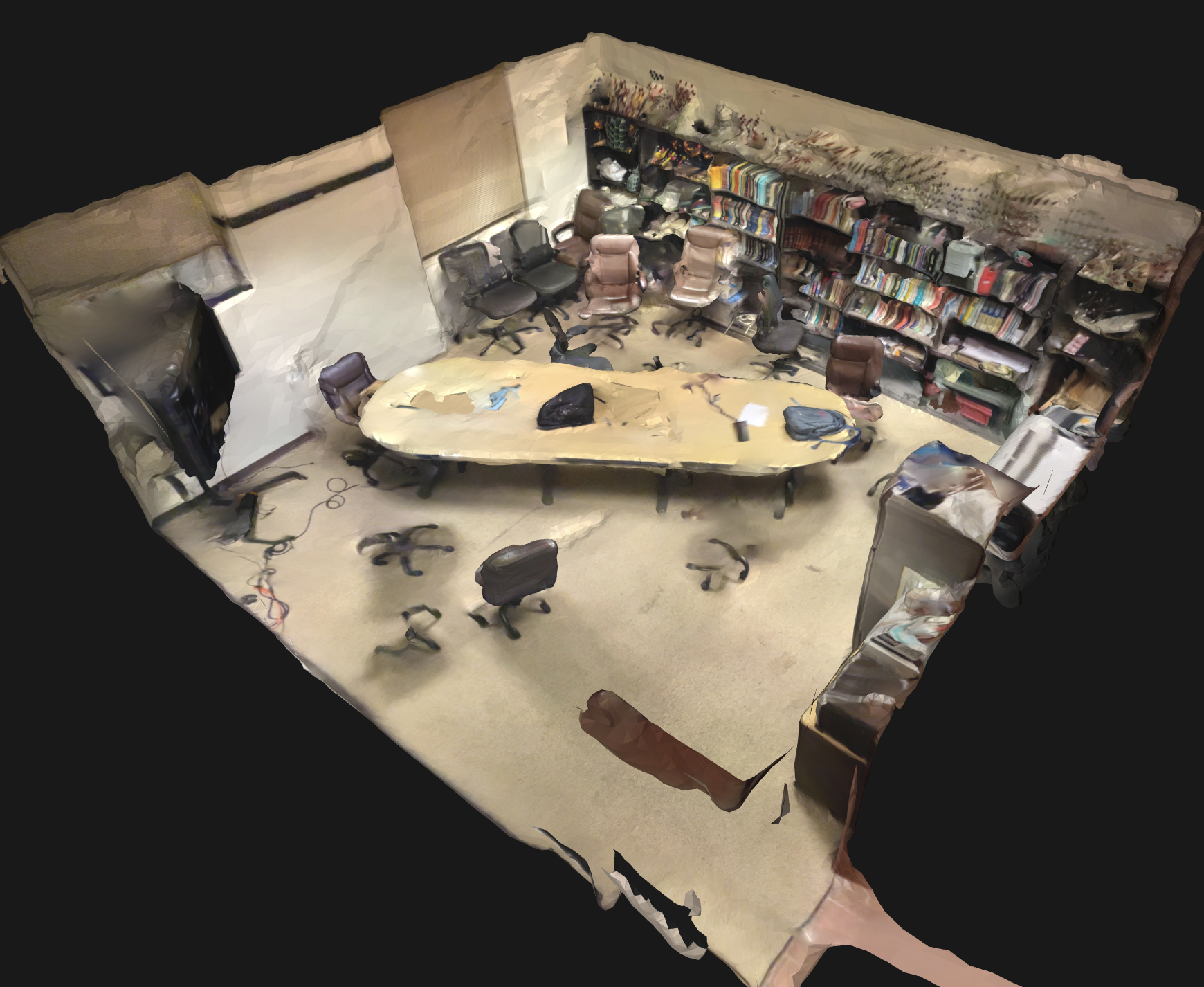}
        \caption{Conference Room.}
        \label{fig:conferenceRoom}
    \end{subfigure}
\caption{3D scans of diverse indoor spaces at our university used to evaluate \systemname{} under varied and challenging conditions.}
\label{fig:mapScreenshots}
\end{figure}

As described in our Evaluation section, we 3D scanned several indoor locations in our university to evaluate the different components of \systemname{}. Figure~\ref{fig:mapScreenshots} shows 3D models of four of our scans---work cubilces, kitchen, a classroom, and a conference room. We scanned spaces with a diverse set of features and objects. This let us evaluate our pipeline under a wide range of challenging conditions. 

\subsection{Samples of images used in evaluation}

Many of our evaluations rely on image data we collected. In this section, we present a sample of images used in our evaluation.

\subsubsection{Masked Localization}

In Section~8.2, we show that masking certain objects (e.g., chairs, keyboards, and people) in our images before localization can improve localization accuracy. To show this, we run localization on two sets of images -- one where the objects are in the same place as in our scans, and the other where certain objects have been moved around.

\begin{figure}
    \centering
    \includegraphics[width=0.95\linewidth]{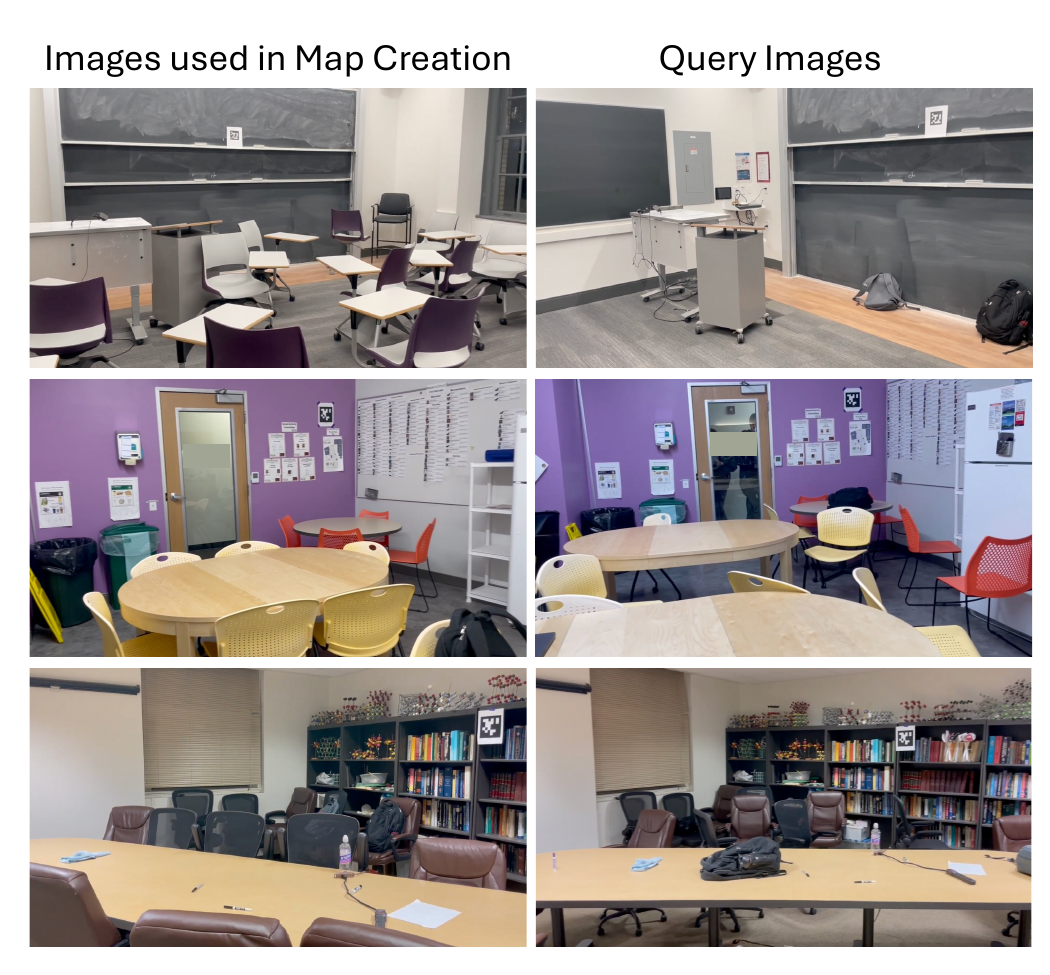}
    \caption{Examples of images used in evaluating masked localization. The images on the left were used in map creation, while the images on the right were used at query time to localize. Notice how the objects such as chairs, backpacks, and tables have been moved or removed in the query images.}
    \label{fig:maskedLocExamples}
\end{figure}

Figure~\ref{fig:maskedLocExamples} shows samples of images used in map creation and query. Once we scan the space, we move around or remove objects such as chairs and tables, and also introduce new objects such as backpacks before capturing query images. These are objects that are typically added, removed or moved in university indoor environments.

\begin{figure}
    \centering
    \includegraphics[width=0.95\linewidth]{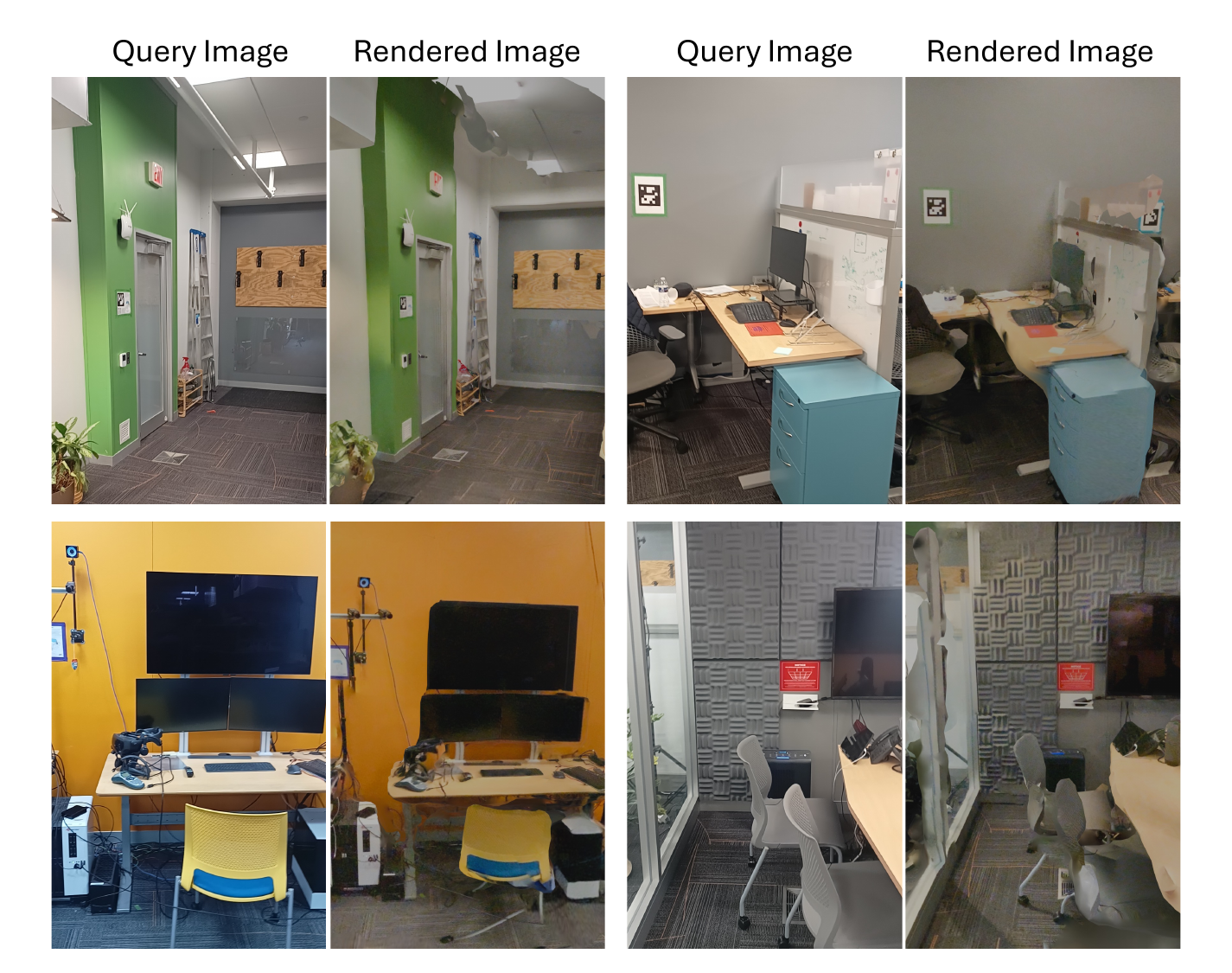}
    \caption{Examples of query images and rendered images generated in pose confidence evaluation. Notice how the images are different in terms of their pixel values but are semantically similar. This results in low similarity scores for pixel-based comparison metrics such as PSNR and SSIM but high scores for similarity metrics based on semantic embedding such as CLIP.}
    \label{fig:poseConfidenceExamples}
\end{figure}

\subsubsection{Pose Confidence}

In Section~8.3, we evaluated our pose confidence component by rendering images corresponding to estimated poses, and comparing the rendered image against the query image. Figure~\ref{fig:poseConfidenceExamples} shows a few examples of rendered and query images. Notice their semantic similarity which is captured by our fine-tuned CLIP model as shown in the paper.

\subsubsection{Place Recognizer}

Place Recognizer component determines if a give query image belongs to the map maintained by a VPS service and was evaluated in Section~8.5. In our evaluation, we considered 5 VPS services that are close and are visually similar to each other. Figure~\ref{fig:placeRecognizerExamples} shows some sample images that were used to fine-tune and query the Place Recognizer module. Notice how some of the rooms, the color scheme and layout are similar to the others.

\begin{figure}
    \centering
    \includegraphics[width=0.95\linewidth]{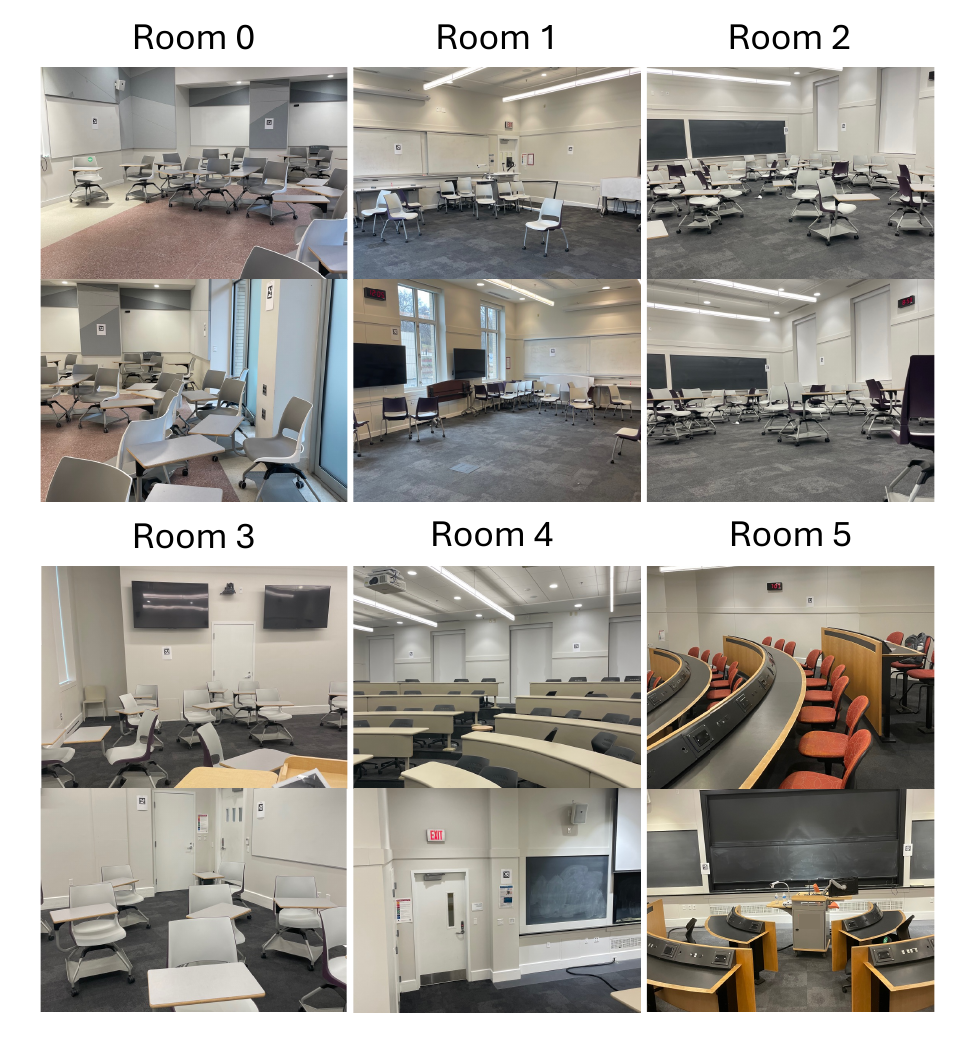}
    \caption{Examples of images from rooms used in the evaluation of the Place Recognizer component. These rooms are close to each other in a building and might be discovered simultaneously. The Place Recognizer component successfully differentiates between rooms even if they look visually similar.}
    \label{fig:placeRecognizerExamples}
\end{figure}

\subsection{Summary of neural network models used}

We used a number of neural network models in various components of \systemname{}. Some of them were fine-tuned for their specific tasks. Table~\ref{tab:modelsUsed} provides a summary of these models.

\begin{table}[]
\resizebox{\columnwidth}{!}{%
\begin{tabular}{|l|l|}
\hline
\textbf{Component}                                               & \textbf{Model}                                                                                                                                                                                                          \\ \hline
Place Recognizer                                                 & \begin{tabular}[c]{@{}l@{}}CLIP (ViT-L/14@336px)~\cite{clip} \\ with fine-tuned 512 X 256 projection head.\end{tabular}                                                                                                 \\ \hline
\begin{tabular}[c]{@{}l@{}}Semantic Image\\ masking\end{tabular} & YOLO11~\cite{yolov11}                                                                                                                                                                                                   \\ \hline
Localization                                                     & \begin{tabular}[c]{@{}l@{}}SuperPoint~\cite{detone2018superpoint} (feature detector)\\ SuperGlue~\cite{sarlin2020superglue} (feature matcher)\\ NetVLAD~\cite{arandjelovic2016netvlad} (global descriptor)\end{tabular} \\ \hline
Pose Confidence                                                  & \begin{tabular}[c]{@{}l@{}}CLIP (ViT-L/14@336px)~\cite{clip} \\ with fine-tuned 512 X 256 projection head\end{tabular}                                                                                                  \\ \hline
\end{tabular}%
}
\caption{Models used in different components of \systemname{}. }
\label{tab:modelsUsed}
\end{table}

\section{Runtime Analysis}

We report the runtime of \systemname{} components in Table~\ref{tab:runtimeAnalysis}. The client-side components were profiled on a low-end Android phone---Motorola Edge 20 Lite (with MediaTek Dimensity 720 processor and 8GB memory). The server-side components were profiled on a server with Intel i9-13900K CPU and NVIDIA GeForce RTX 4090 GPU. 

\systemname{} runs the VPS Discoverer, Selector and Stitcher modules on the client. These are light-weight processes that do not affect the application performance. VPS Discoverer makes simple queries to a remote database and does not significantly use the client's compute resources. The Stitcher and Selector involve small 4X4 matrix multiplications. All of these modules run under 5~ms and do not interfere with other parts of the application. Furthermore, these simple algorithms can run on resource constrained devices such as thin AR glasses.

\systemname{} VPS servers run some resource intensive algorithms such as Place Recognizer, localization against large maps, and Pose Confidence. As the client makes asynchronous or non-blocking requests to VPS services, the runtime of server-side components does not directly affect the user experience. 

Furthermore, only the Pose Estimation component is mandatory on a VPS service. All other components can be turned off or swapped out with a computationally less intensive process, if required. As our VPS Service is modularized into independent components, it allows making trade-offs between performance and accuracy. Some examples of such trade-off decisions are:

\begin{itemize}
    \item Pose Confidence, which is the most resource intensive component in our implementation, can be turned off as the clients have their own way of calculating confidences in the VPS Selector module. It can also be swapped with a less resource intensive technique some of which are discussed in the paper (e.g., percentage of inliers).
    \item The Place Recognizer component can be turned off if the VPS service maintainer recognizes that most localization requests pass onto the other stages and Place Recognizer is not reducing resource usage.
    \item Semantic masking can be turned off in environments where no improvement in localization accuracy is observed by masking.
\end{itemize}

\begin{table}[]
\resizebox{\columnwidth}{!}{%
\begin{tabular}{|r|l|l|}
\hline
\multicolumn{1}{|l|}{\textbf{Component}}                                 & \textbf{\begin{tabular}[c]{@{}l@{}}Mean\\ (ms)\end{tabular}} & \textbf{\begin{tabular}[c]{@{}l@{}}Standard Deviation \\ (ms)\end{tabular}} \\ \hline
\rowcolor[HTML]{C8E4F3} 
\multicolumn{1}{|l|}{\cellcolor[HTML]{C8E4F3}\textbf{VPS Selector}}      & \textbf{1.43}                                                & \textbf{0.09}                                                               \\ \hline
\rowcolor[HTML]{C8E4F3} 
\multicolumn{1}{|l|}{\cellcolor[HTML]{C8E4F3}\textbf{VPS Stitcher}}      & \textbf{0.03}                                                & \textbf{0.01}                                                               \\ \hline
\rowcolor[HTML]{ECD0EC} 
\multicolumn{1}{|l|}{\cellcolor[HTML]{ECD0EC}\textbf{Place Recognizer}}  & \textbf{25.48}                                               & \textbf{1.47}                                                               \\
\rowcolor[HTML]{ECD0EC} 
CLIP Inference                                                           & 25.35                                                        & 1.48                                                                        \\
\rowcolor[HTML]{ECD0EC} 
Database Query                                                           & 0.13                                                         & 0.01                                                                        \\ \hline
\rowcolor[HTML]{ECD0EC} 
\multicolumn{1}{|l|}{\cellcolor[HTML]{ECD0EC}\textbf{Semantic Masking}}  & \textbf{9.04}                                                & \textbf{1.83}                                                               \\ \hline
\rowcolor[HTML]{ECD0EC} 
\multicolumn{1}{|l|}{\cellcolor[HTML]{ECD0EC}\textbf{Pose Estimation}}   & \textbf{289.21}                                              & \textbf{67.46}                                                              \\
\rowcolor[HTML]{ECD0EC} 
\begin{tabular}[c]{@{}r@{}}Feature detection\\ (Superpoint)\end{tabular} & 9.16                                                         & 6.15                                                                        \\
\rowcolor[HTML]{ECD0EC} 
\begin{tabular}[c]{@{}r@{}}Global Descriptor\\ (NetVLAD)\end{tabular}    & 14.46                                                        & 1.72                                                                        \\
\rowcolor[HTML]{ECD0EC} 
Candidate Matches                                                        & 8.27                                                         & 0.04                                                                        \\
\rowcolor[HTML]{ECD0EC} 
\begin{tabular}[c]{@{}r@{}}Feature Matching\\ (SuperGlue)\end{tabular}   & 188.86                                                       & 56.21                                                                       \\
\rowcolor[HTML]{ECD0EC} 
Get Pose from SfM                                                        & 68.47                                                        & 38.82                                                                       \\ \hline
\rowcolor[HTML]{ECD0EC} 
\multicolumn{1}{|l|}{\cellcolor[HTML]{ECD0EC}\textbf{Pose Confidence}}   & \textbf{634.00}                                              & \textbf{8.04}                                                               \\
\rowcolor[HTML]{ECD0EC} 
Blender Render                                                           & 562.60                                                       & 6.49                                                                        \\
\rowcolor[HTML]{ECD0EC} 
CLIP Inference                                                           & 71.41                                                        & 3.42                                                                        \\ \hline
\end{tabular}%
}
\caption{Runtime anlysis of \textcolor[HTML]{3EB4F3}{Client} and \textcolor[HTML]{EC65EC}{Server} components in \systemname{}. Notice that the components running on the client (VPS Selector and VPS Stitcher) are lightweight and can run on resource constrained clients.}
\label{tab:runtimeAnalysis}
\end{table}
\section{Discussion}

\subsection{Integration with existing systems}

\systemname{} has the potential to become a ubiquitous localization backend for both indoor and outdoor environments by integrating outdoor VPS services like Google Geospatial API and Niantic Lightship VPS. Such services would act as independent \textit{VPS Services} on \systemname{} and would be discovered whenever the client device is outdoors. There are two possible solutions---short-term and long-term---to integrating existing VPS services with \systemname{}.

\begin{itemize}
    \item \textbf{Long term -- Standardized API implementation}: Currently, existing VPS services are tightly coupled with their Software Development Kits (SDKs). Since there is no standardization around VPS services and their application-facing interfaces yet, SDKs from different companies, such as Google and Niantic, are not interoperable to the best of our knowledge. We envision that in the long-term as AR applications mature, VPS providers would adhere to a standard and implement a common set of APIs that can be used by \systemname{}. We believe that the VPS Service interfaces that we discuss in the paper could help with such standardizations in the future.

    \item \textbf{Short term -- Proxy Servers}: In the short term, we could set up proxy servers that would expose interfaces similar to other VPS Services on \systemname{}. These servers would forward localization requests from applications to existing VPS services.
\end{itemize}

\subsection{Scalability}

\systemname{}'s scalability derives from its federated architecture, where independent organizations can deploy VPS services without requiring coordination or central infrastructure. This enables the system to scale horizontally as new services are added. Traditional scalability metrics (e.g., centralized throughput tests, stress tests etc) are less meaningful in this context. Instead, we demonstrate practical feasibility via deployment of 30 VPS services across varied indoor environments, reflecting the system’s capacity for distributed growth. However, true validation of large-scale scalability will come only from broader real-world adoption of \systemname{}.

\end{document}